\documentclass[10pt,twocolumn,letterpaper]{article}

\usepackage[pagenumbers]{lib/iccv} 

\definecolor{iccvblue}{rgb}{0.21,0.49,0.74}
\usepackage[pagebackref,breaklinks,colorlinks,allcolors=iccvblue]{hyperref}
\usepackage[utf8]{inputenc} 
\usepackage[T1]{fontenc}    
\usepackage{url}            
\usepackage{booktabs}       
\usepackage{amsfonts}       
\usepackage{nicefrac}       
\usepackage{xcolor}         
\usepackage{graphicx}
\usepackage{amsmath}
\usepackage{amssymb}
\usepackage{bm}
\usepackage{makecell}
\usepackage{multirow}
\usepackage{xspace}
\usepackage{overpic}
\usepackage{colortbl}
\usepackage{wrapfig}
\usepackage{lipsum} 
\usepackage{threeparttable}
\usepackage[normalem]{ulem}

\definecolor{MyDarkRed}{rgb}{0.46, 0.16, 0.16}
\definecolor{MyDarkBlue}{rgb}{0.16, 0.16, 0.66}
\definecolor{MyPink}{rgb}{1.0, 0.702, 0.729} 
\definecolor{MyPeach}{rgb}{1.0, 0.875, 0.729} 
\definecolor{MyLightYellow}{rgb}{1.0, 1.0, 0.729} 
\definecolor{MyLightGreen}{rgb}{0.729, 1.0, 0.788} 
\definecolor{MyLightBlue}{rgb}{0.729, 0.882, 1.0} 

\newcommand{\pa}{PolarAnything\xspace}

\newcommand{\our}{PolarAnything\xspace}

\newcommand{\pandora}{PANDORA~\cite{pandora}\xspace}
\newcommand{\nersp}{NeRSP~\cite{nersp}\xspace}

\newcommand{\nero}{NeRO~\cite{nero}\xspace}

\newcommand{\mitsuba}{Mitsuba~\cite{mitsuba2}\xspace}

\newcommand{\deepsfp}{DeepSfP~\cite{ba19deepsfp}}
\newcommand{\restormer}{Restormer~\cite{restormer}\xspace}
\newcommand{\diligent}{DiliGenT~\cite{diligent}\xspace}

\newcommand{\pisr}{PISR~\cite{pisr}\xspace}

\newcommand{\Tref}[1]{Table~\ref{#1}}
\newcommand{\eref}[1]{Eq.~\eqref{#1}}

\newcommand{\fref}[1]{Fig.\ \ref{#1}}
\newcommand{\Fref}[1]{Figure~\ref{#1}}

\usepackage{xcolor}
\newcounter{todos}
\AtEndDocument{\ifnum\value{todos}>0 \PackageWarning{TODOS}{There are \arabic{todos} todos left in this paper! Fix them before submitting the paper!} \fi}

\newcommand{\V}[1]{\ensuremath{\mathbf{#1}}}

\makeatletter
\DeclareRobustCommand\onedot{\futurelet\@let@token\@onedot}
\def\@onedot{\ifx\@let@token.\else.\null\fi\xspace}
\def\eg{\emph{e.g}\onedot} 
\def\ie{\emph{i.e}\onedot}

\def\etal{\emph{et al}\onedot}
\makeatother

\def\paperID{4060} 
\def\confName{ICCV}
\def\confYear{2025}

\title{PolarAnything: Diffusion-based Polarimetric Image Synthesis}

\author{Kailong Zhang$^{1\dagger}$~~~Youwei Lyu$^{1\dagger}$~~~Heng Guo$^{1,2}\thanks{\,corresponding author.\,\,\, $\dagger$ equally contributed authors.}$~~~Si Li$^{1}$~~~Zhanyu Ma$^{1}$~~~Boxin Shi$^{3,4}$\\
\small{$^1$Beijing University of Posts and Telecommunications},  \small{$^2$Xiong'an Aerospace Information Research Institute}\\
\small{$^3$State Key Laboratory of Multimedia Information Processing, School of Computer Science, Peking University}\\
\small{$^4$National Engineering Research Center of Visual Technology, School of Computer Science, Peking University}\\
\small{\texttt{\{zhangkailong, youweilv, guoheng, lisi, mazhanyu\}@bupt.edu.cn~~~shiboxin@pku.edu.cn}}
}

\begin{document}
\maketitle
\begin{abstract}
Polarization images facilitate image enhancement and 3D reconstruction tasks, but the limited accessibility of polarization cameras hinders their broader application. 
This gap drives the need for synthesizing photorealistic polarization images.
The existing polarization simulator Mitsuba relies on a parametric polarization image formation model and requires extensive 3D assets covering shape and PBR materials, preventing it from generating large-scale photorealistic images. To address this problem, we propose \pa, capable of synthesizing polarization images from a single RGB input with both photorealism and physical accuracy, eliminating the dependency on 3D asset collections. 
Drawing inspiration from the zero-shot performance of pretrained diffusion models, we introduce a diffusion-based generative framework with an effective representation strategy that preserves the fidelity of polarization properties. Experiments show that our model generates high-quality polarization images and supports downstream tasks like shape from polarization. The project could be found in \href{https://flzt11.github.io/PA\_project/}{https://flzt11.github.io/PA\_project/}.
\end{abstract}    
\section{Introduction}
\label{sec:intro}

Polarization-based computer vision tasks, including shape-from-polarization~\cite{atkinson2006sfp,ba19deepsfp,lyu2023distltsfp}, image dehazing~\cite{zhou2021polarDehaze}, and reflection removal~\cite{kong14polarRR,reflectnet,lei20polarRR,lyu23polarRR}, leverage additional polarization cues to enhance RGB-based methods. 
Different from RGB cameras commonly equipped on mobile phones, polarization cameras are expensive and inaccessible to most users, limiting the application range of polarization-based vision systems. On the other hand, diverse and large-scale polarization image-based datasets with ground truth (GT) labels such as depth, surface normals, and semantic segmentation masks rarely exist, but these datasets are important for improving the performance of the corresponding learning-based computer vision tasks. In contrast, such datasets based on RGB images are widely available.
Therefore, it is desired to develop a polarization image simulator taking as input RGB images to easily create polarization datasets and make the polarization-based vision algorithms accessible to users even without specialized polarization cameras.

\begin{figure}
    \includegraphics[width=\linewidth]{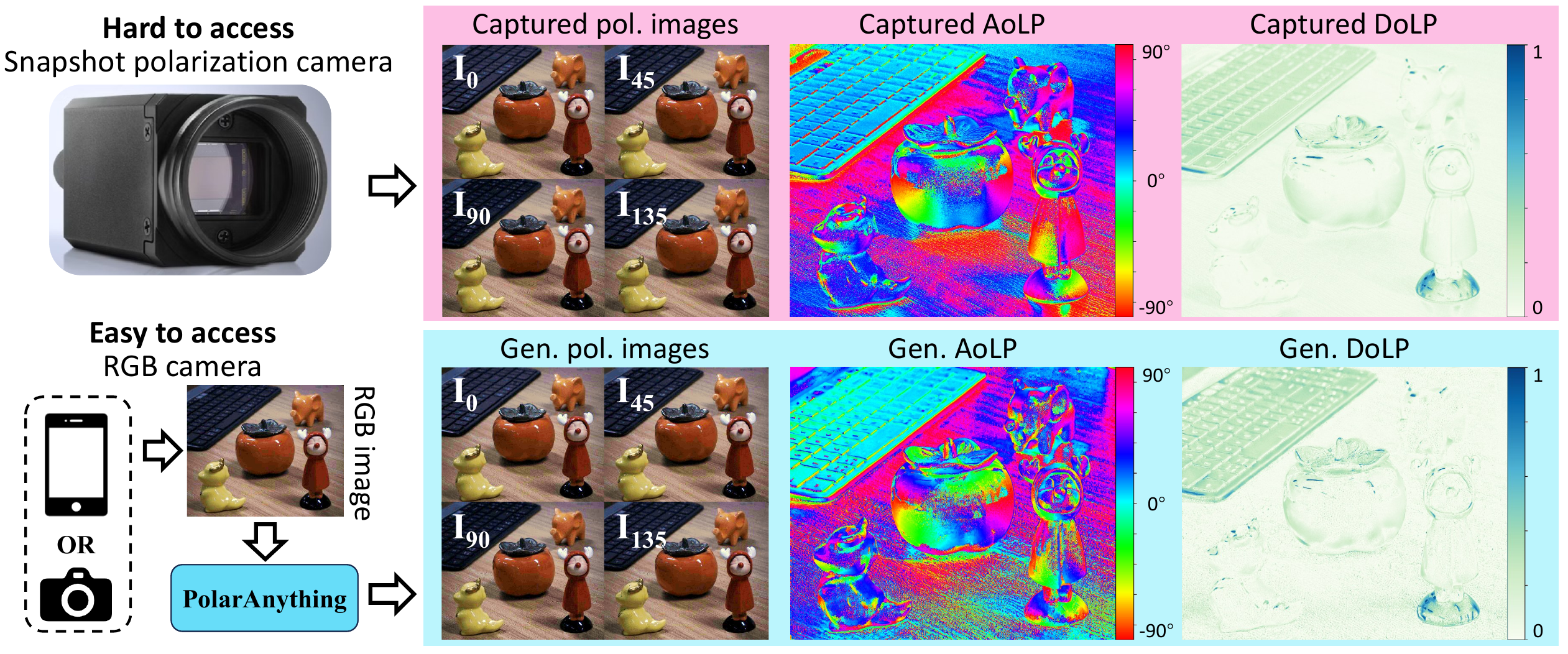}
    \caption{\pa generates polarization images covering diverse shapes and materials from a single image captured by an accessible RGB camera while preserving the polarization properties on both AoLP and DoLP. The top and bottom rows show the polarization images captured by a snapshot polarization camera and the ones generated by \pa, respectively.}
    \label{fig:teaser}
    \vspace{-5mm}
\end{figure}

Existing polarization image simulators are either based on collected measured polarized Bidirectional Reflectance Distribution Function~(pBRDF)~\cite{Baek20polarReflectance} or a parametric polarized reflectance model~\cite{baek2018pbrdf,kondo20pBRDF,Ichikawa23pFBRDF}. Due to the scarcity of polarization datasets, data-based simulators using measured pBRDFs are inherently limited in representing diverse real-world scenes. On the other hand, a simulator based on a parametric pBRDF model, such as \mitsuba, still has a gap in synthesizing polarization properties of real-world scenes due to its reflectance assumption, as demonstrated by fitting the measured pBRDF in \cite{Baek20polarReflectance}. Moreover, Mitsuba requires 3D assets including both surface geometry and paired physically-based material maps such as roughness and albedo to render polarization images. Compared to RGB images, 3D assets are hard to collect, and their amount is much smaller than that of RGB images. Besides, even with plenty of 3D assets, selecting and seamlessly integrating them into a scene for scene-level polarization image synthesis is even more challenging. To summarize, current polarization simulators still face considerable difficulties in generating large-scale photorealistic polarization images.

To handle these problems, we propose \pa, a polarization image simulator taking a single RGB image as input, capable of handling {diverse shape and materials}, as shown in \fref{fig:teaser}.
RGB images are easier to acquire against 3D assets and there are plenty of RGB-based datasets with labeled ground truth~(\eg, depth, surface normal, segmentation) for a broad range of vision tasks. Furthermore, objects in daily RGB images are naturally distributed, making simulated polarization images photorealistic. To eliminate assumptions inherent in parametric pBRDF models, we propose a diffusion-based model for polarization image generation. Specifically, this model conditions an RGB image to perform diffusion on the encoded Angle of Linear Polarization~(AoLP) and Degree of Linear Polarization~(DoLP). Compared to simply conducting diffusion on the polarization images under different polarization angles, we show such a design can better preserve polarization information. 

Based on \pa, we show existing RGB-based datasets such as Stanford-ORB~\cite{kuang2023stanfordorb} can be transferred to their polarimetric version. Specifically, we create Polar-Stanford-ORB with 300 paired polarimetric images and their corresponding GT surface normal maps, benefiting the training of learning-based polarimetric vision tasks such as DeepSfP~\cite{ba19deepsfp} requiring labeled GT.

To summarize, the contributions are threefold:
\begin{itemize}[]
    \item We propose \pa, the first polarization image simulator based on RGB image as input, reducing the effort required to obtain polarization images;
    \item We show that  finetuning Stable Diffusion on the AoLP and DoLP instead of the polarization images can better preserve the polarization information;
    \item We create PolarStanford-ORB based on our photorealistic polarization image synthesis and demonstrate the effectiveness of \pa on downstream polarization-based computer vision tasks.
\end{itemize}

\begin{figure*}
    \includegraphics[width=\linewidth]{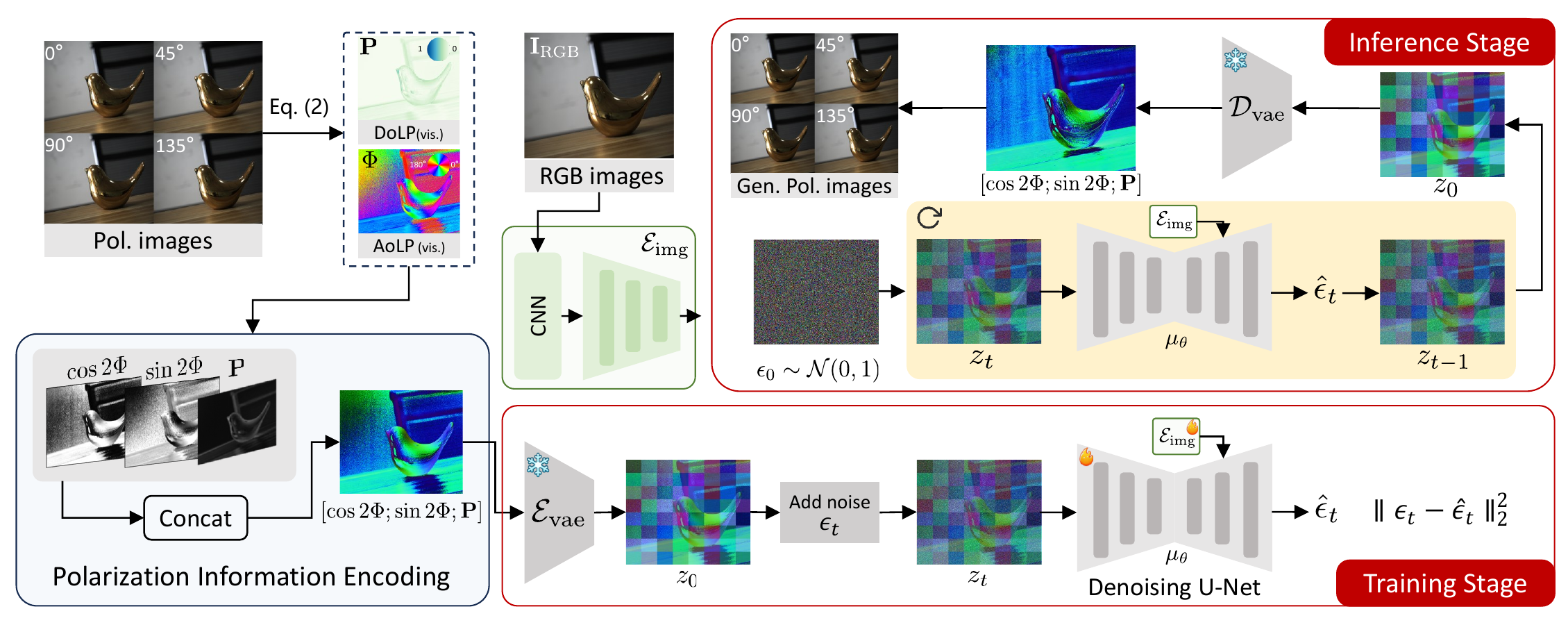}
    \vspace{-1.5em}
    \caption{Framework of PolarAnything. In the training stage, the image encoder takes as input the RGB image and extracts the conditional feature for prompting U-Net denoising. During inference, PolarAnything is fed with an RGB image and generates high-quality AoLP and DoLP via steps of denoising.}
    \label{fig:pipeline}
    \vspace{-3mm}
\end{figure*}
\section{Related Works}
\subsection{Polarization Applications and Acquisition}
\paragraph{Polarization applications} By providing additional measurements, polarization images can enhance a wide range of downstream vision tasks in various ways. Polarization encodes detailed geometry and reflectance information of object surfaces and is well studied for surface normal recovery \cite{miyazaki03sfp,atkinson2006sfp,ba19deepsfp,shapeFromSky,sfp_wild,chen22ppa,lyu2023distltsfp}, depth estimation~\cite{kadambi2017depth,smith2018height,DPSNet,ikemura24polarDepth}, 3D reconstruction~\cite{mvir,pandora,nersp,neisf,chen24PISR}, and reflectance decompositions~\cite{Ghosh10circPolarReflectance,baek2018pbrdf,Baek20polarReflectance}, further imposing geometric constraints on 6D pose predictions~\cite{gao22polarPoseEst}. As an important light property, polarization status reflects the light propagation process, such as scattering in air and refraction in the semi-reflector, and provides unique cues for solving dehazing~\cite{zhou2021polarDehaze}, reflection removal~\cite{kong14polarRR,reflectnet,lei20polarRR,lyu23polarRR}, and transparent object segmentation~\cite{polarTransSeg,mei22glassSeg} problems. Light polarization characterizes oscillation magnitudes along two orthogonal orientations on the 2D frame, which inspires researchers to utilize polarized light for designing the structure light system~\cite{polStructureLt} and alpha matte extraction setup~\cite{polarAlpha}. 
\paragraph{Polarization image capture and synthesis}Polarization image capture requires a specialized polarization camera or manually rotated polarizer filters, and the laboursome acquisition process restricts broader applications of polarization images. Mitsuba 2~\cite{mitsuba2} is a retargetable forward and inverse renderer aiming to formulate the light transport process, which enables polarization image rendering and data synthesis. However, it is necessary to carefully configure light conditions and use a large amount of tailored 3D models and Physically Based Rendering (PBR) materials in rendering, making the large-scale production of high-quality polarization data costly. \pa aims to take as input the most common RGB images and produce polarization images based on a diffusion model at a low cost, further facilitating polarization research for the community.
\subsection{Image Synthesis with Diffusion Models}
With the advent of deep learning, data-driven approaches have been the widely adopted tool for data synthesis. Given large-scale datasets for supervision, UNet-based convolutional neural networks could generate 4D RGBD light-field images~\cite{lfgen17}, and dual-pixel images~\cite{dpGen}, and simulate motion blur~\cite{motionBlurGen}. Adversarial generative network (GAN)~\cite{gan} is shown to be a powerful generative method, while diffusion models~\cite{diffusion,ddpm,ddim} demonstrate superior image synthesis capabilities. 
Denoising Diffusion Probabilistic Models (DDPMs)~\cite{ddpm} have become an emerging research spot of generative models, which present impressive high-quality image synthesis results. DDPMs learn to invert a diffusion process that incrementally corrupts images with Gaussian noise, enabling them to generate samples from the data distribution by applying the reverse process to random noise. Conditional diffusion models extend DDPMs by incorporating additional information, such as text~\cite{text2img} and images~\cite{img2imgDiff}. In the field of text-based image generation, Stable Diffusion~\cite{sd} is based on a latent diffusion model trained on extensively large-scale datasets~\cite{laion} and achieves unprecedented image synthesis quality. 
With distilled knowledge from vast data, the strong diffusion priors could help to produce surprising zero-shot performance. Several studies have investigated methods to leverage the powerful pretrained model.
ControlNet~\cite{controlnet} proposes a general finetune framework to integrate additional conditions with pre-trained diffusion models such as Stable Diffusion~\cite{sd} to accomplish more customized tasks. Recent works, StableNormal~\cite{stablenormal}, GeoWizard~\cite{fu2024geowizard}, and Marigold~\cite{marigold}, propose to use diffusion models as geometric information predictors. Gao~\etal~\cite{lfdiff} design a position-aware warping scheme providing an initial light field pattern, which conditions the diffusion model to generate light field images. StereoDiffusion~\cite{stereoDiff} introduces the disparity map to guide the stereo pixel shift operation into the diffusion process, which predicts stereo images without finetuning the original model. Motivated by diffusion models generating remarkable results in discriminative tasks, we propose \pa to restore polarization properties based on powerful pretrained priors, achieving physically reliable polarization prediction.
\section{Method}
\our is a generalizable latent diffusion model for synthesizing polarimetric observations under arbitrary polarization angles from an RGB image. We achieve this goal by creating a high-quality polarization image dataset and finetuning a pretrained image-conditioned generative model on the dataset. \Fref{fig:pipeline} displays the overall architecture of \our. 

\subsection{Preliminaries on Polarization}\label{sec:3.1}
Polarization images are acquired with polarizer filtering at different polarizer angles. The intensity of the light passing through a polarizer with an angle of $\Theta$ is given by:
\begin{equation}\label{eq:1}
    \mathbf{I}_\Theta = \frac{\mathbf{I}_\text{RGB}}{2}(1+\mathbf{P} \cos{(2\Theta-2\Phi)}),
\end{equation}
in which $\mathbf{I}_\text{RGB}$ denotes incoming light intensities before the polarizer. DoLP $\mathbf{P}$ denotes the strength ratio of linearly polarized light to total incident light. $\Phi$ denotes AoLP, \ie, the oscillation orientation of the polarized component. 
With a snapshot polarization camera, four polarization images, \ie, $\mathbf{I}_{0^\circ},\mathbf{I}_{45^\circ},\mathbf{I}_{90^\circ},\mathbf{I}_{135^\circ}$, can be obtained with a single shot, and the polarization properties AoLP and DoLP can be computed as follows: 
\begin{equation}\label{eq:2}
\begin{split}
    &\Phi = \frac{1}{2}atan2{\frac{\mathbf{I}_{45^\circ}-\mathbf{I}_{135^\circ}}{\mathbf{I}_{0^\circ}-\mathbf{I}_{90^\circ}}}, 
    \\
    &\mathbf{P} = \frac{\sqrt{(\mathbf{I}_{0^\circ}-\mathbf{I}_{90^\circ})^2+(\mathbf{I}_{45^\circ}-\mathbf{I}_{135^\circ})^2}}{(\mathbf{I}_{0^\circ}+\mathbf{I}_{45^\circ}+\mathbf{I}_{90^\circ}+\mathbf{I}_{135^\circ})/2}.
\end{split}
\end{equation}

\subsection{PolarAnything}\label{sec:3.2}

\paragraph{Polarized information encoding} For polarization image generation, one option is to denoise polarization images conditioned on specific polarizer angles. However, we observe that the polarization properties computed from the image are damaged, which may be attributed to the absence of physical polarization constraint during the denoising steps. To generate physically plausible polarization properties, we propose to estimate corresponding AoLP and DoLP maps and then simulate polarization images under arbitrary polarizer angles. Considering the $\pi$-periodicity of AoLP, we encode AoLP in the sinusoidal form as $(\cos{2\Phi}, \sin{2\Phi})$, which is also a continuous representation facilitating network learning~\cite{continueRepre}. Consistent with the value range of encoded AoLP and the VAE, we normalize DoLP within $[-1, 1]$, and then concatenate the encoded AoLP with DoLP as diffusion output: $[\cos{2\Phi}; \sin{2\Phi}; \mathbf{P}]$. The image of encoded AoLP and DoLP is visualized in \fref{fig:pipeline}. Given an RGB image $\mathbf{I}_\text{RGB}$ and its AoLP and DoLP, polarization images of arbitrary polarizer angles can be obtained by \eref{eq:1}.

\paragraph{Diffusion-based polarization generator} Pre-trained diffusion models have shown impressive zero-shot performance in text-based image generation and discriminative tasks.
A Denoising Diffusion Probabilistic Model (DDPM) depicts data distributions by progressively transforming noisy samples via reverse diffusion processes $z_{t-1}=\beta_tz_t-\mu_\theta(z_t, t)+\epsilon_t$, where $\beta_t$ is a variance schedule, $\epsilon_t$ denotes the noise added at timestep $t$, and the model learns to iteratively denoise $z_t$ via U-Net $\mu_\theta$ parameterized by $\theta$. 

We propose \our to generate polarization properties from a sole RGB image $\mathbf{I}_\text{RGB}$ by a latent denoising diffusion model, which models the conditional distribution $p(\cos{2\Phi};\sin{2\Phi};\mathbf{P}|\mathbf{I}_\text{RGB})$. The overview of \our is shown in \fref{fig:pipeline}. \our is based on the pretrained large diffusion model Stable Diffusion v1.5~\cite{sd}, which learns representative and generalizable image priors from LAION-5B~\cite{laion}. The model consists of an image encoder $\mathcal{E}_\text{img}$, a pretrained VAE encoder $\mathcal{E}_\text{vae}$, a denoising U-Net $\mu_\theta$, and a pretrained VAE decoder $\mathcal{D}_\text{vae}$. 
We obtain the conditional signal by encoding the input RGB image with the condition encoder $\mathcal{E}_\text{img}$. The condition encoder comprises two cascaded modules: the first is an RGB image feature extractor, which consists of several convolutional layers with SiLU activations; the latter feature encoder shares the same structure as that of the denoising U-Net encoder. The condition encoder $\mathcal{E}_\text{img}$ generates hierarchical guidance features for feature fusion in the decoder of denoising U-Net like ControlNet~\cite{controlnet}. 
The VAE encoder $\mathcal{E}_\text{vae}$ and decoder $\mathcal{D}_\text{vae}$ conduct the conversion between the encoded AoLP and DoLP map and its latent code $z_{t}$. The weights of the VAE encoder and decoder are fixed in the finetuning stage. The U-Net $\mu_\theta$ is conditioned on the hierarchical guidance features
$\mathcal{E}_\text{img}(\mathbf{I}_\text{RGB})$ to denoise polarization properties in the latent space. We finetune \our with the following objective:
\begin{equation}
    \min_\theta\mathbb{E}_{x\sim\mathcal{E}_\text{vae},t,\epsilon\sim\mathcal{N}(0,1)}||\epsilon_t-\mu_\theta(z_t,t,c,\mathcal{E}_\text{img}(\mathbf{I}_\text{RGB}))||^2_2,
\end{equation}
in which $c$ is the text prompt embedding generated by CLIP~\cite{clip}. 

At the inference stage, the latent code of polarization property maps is initialized as standard Gaussian noise, which is iteratively denoised with the same schedule as finetuning.  

\subsection{Real-world Polarization Dataset}\label{sec:3.3}
To generate more realistic and physically plausible polarization images, we train \our solely with real captured images.
Collecting polarization images requires manual rotation of a lens-mounted polarizer or a specialized polarization camera, which makes large-scale dataset creation time-consuming. Current real-world polarization image datasets are insufficient to support \our training. 1) Small amount of high-quality data: Morimatsu~\etal~\cite{poldemosaic20} and Qiu~\etal~\cite{qiu21poldemosaic} provide only 40 and 38 sets of polarization images, respectively; 2) large amount of data captured in limited scenes: HAMMER~\cite{hammer}, HouseCat6D~\cite{housecat6d}, and PPP~\cite{gao22polarPoseEst} contain 930, 2358, 5000 polarization images, respectively, but are captured in similar indoor scenes mainly for 6D pose estimation.

To facilitate \pa training and testing, we use a polarization camera~\cite{polarcam} to create an RGB polarization image dataset featuring various shapes and reflectances under divergent light conditions. Our real-world dataset comprises 1,148 high-quality polarization images, each with a resolution of 1224$\times$1024. Over 100 different objects were included in the acquisition, spanning a broad spectrum of categories: transparent/opaque, conductive/dielectric, and diffuse/specular surfaces. Given that illumination can influence polarization properties~\cite{shapeFromSky}, we captured images under 19 varied lighting environments, including 8 outdoor scenes and 11 indoor scenes. An overview of our camera rig and part of the objects is provided in the supplementary material.
We combine data from Morimatsu~\etal~\cite{poldemosaic20} with 1,115 images from our real-world dataset to form the \pa training dataset, while the remaining 33 images are reserved for evaluating generation quality.
\section{Experiment}\label{sec:exp}

\begin{figure*}[t]
    \centering
    \includegraphics[width=0.93\textwidth]{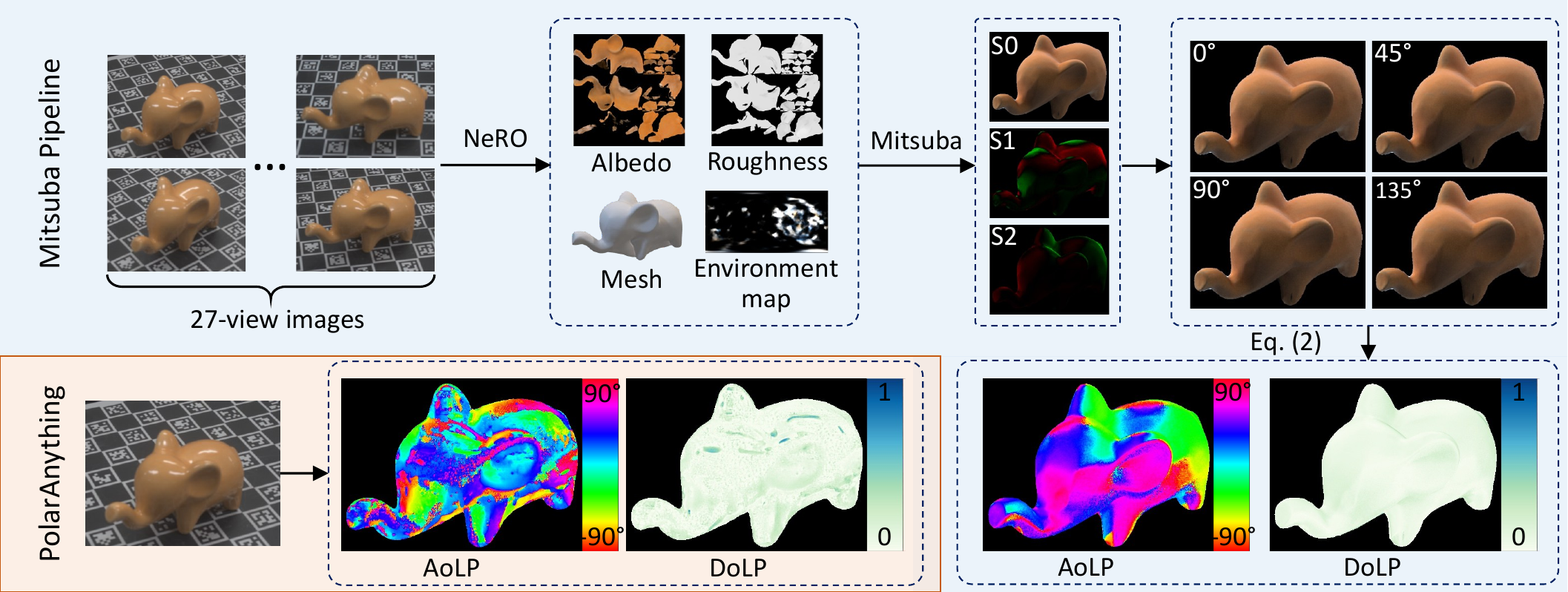}
    \caption{Comparison between Mitsuba and \pa pipelines for generating polarization properties. Multiview images of {\sc Elephant} are captured to generate the mesh, environment light, and PBR parameter maps for rendering polarization images with \mitsuba, while \pa only needs one RGB image for polarization properties generation.}
    \label{fig:pipeline_cmp_mit_pa}
    \vspace{-4mm}
\end{figure*}

\subsection{Implementation Details}

We finetuned the pretrained Stable Diffusion V1.5 backbone using the AdamW~\cite{adamw} optimizer with parameters $\beta_1$=0.9, $\beta_2$=0.999, and a weight decay of 0.001. We set a fixed learning rate of 4$\times$10\textsuperscript{-5} and a batch size of 16 during finetuning. We finetuned the model for 600 steps on 8 NVIDIA A100 cards, which took about 10h. Different from ControlNet~\cite{controlnet}, all the weights of denoising U-Net are trainable during finetuning, which is experimentally proved to be more effective.
For data augmentation, we randomly crop the original 1224$\times$1024 polarization image into 512$\times$512 patches. Since polarization images captured by ~\cite{polarcam} consist of 3 channels(R-G-B), we first convert them to grayscale and then compute DoLP and AoLP using \eref{eq:2} as the ground truth. 

\subsection{Comparison on Polarization Image Synthesis}\label{sec:4.2}
\paragraph{Evaluation metric} To measure the difference between GT and generated polarization images $\{\V{I}_{0}, \V{I}_{45}, \V{I}_{90}, \V{I}_{135}\}$, we use PSNR and SSIM as evaluation metrics. To evaluate generated polarization properties, mean absolute error~(MAbsE) and mean angular error~(MAngE) are adopted to assess generated DoLP and AoLP, respectively.

Due to the $\pi$-periodicity of AoLP, we formulate the angular error between GT and synthesized AoLP~($\Phi$, $\hat{\Phi}$) as

\begin{equation}
	{\rm AngE}(\Phi, \hat{\Phi}) = \min\left\{\left|\hat{\Phi}+k\pi - \Phi\right|,k=-1,0,1\right\}.
\end{equation}

\begin{figure}[t]
    \centering
    \newcommand{\imagesize}{0.120}
\renewcommand{\arraystretch}{0.2}
\resizebox{\linewidth}{!}{
\setlength{\tabcolsep}{1pt}{
\begin{tabular}{cc@{}c@{}cc}
    Image & GT & Mitsuba & PolarAnything &  
    \\
    \begin{subfigure}[c]{\imagesize\textwidth}
        \centering
        \includegraphics[width=\textwidth]{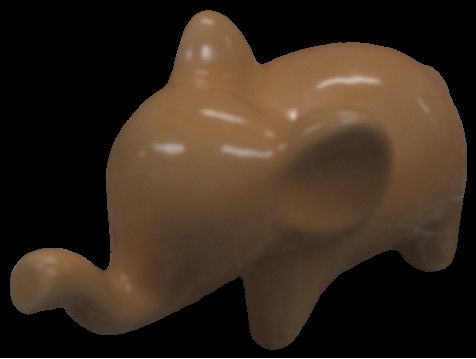}
    \end{subfigure} &
    \begin{subfigure}[c]{\imagesize\textwidth}
        \centering
        \includegraphics[width=\textwidth]{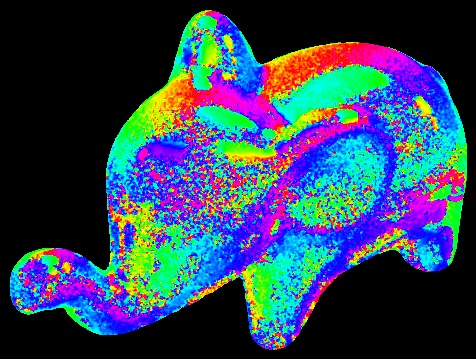}
    \end{subfigure} &
    \begin{subfigure}[c]{\imagesize\textwidth}
        \centering
        \includegraphics[width=\textwidth]{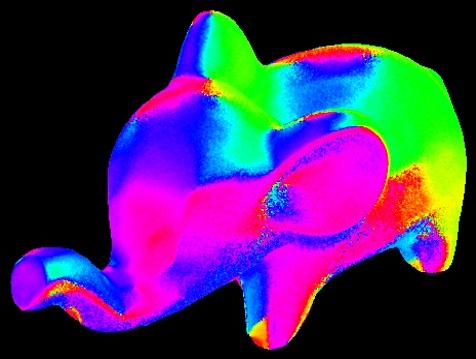}
    \end{subfigure} &
    \begin{subfigure}[c]{\imagesize\textwidth}
        \centering
        \includegraphics[width=\textwidth]{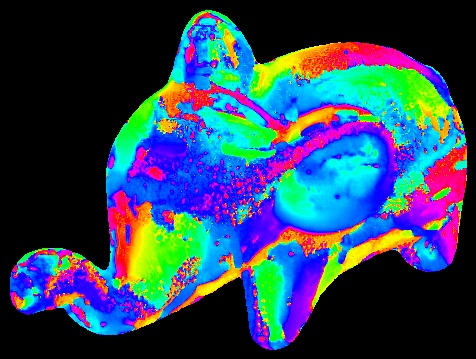}
    \end{subfigure} &
    \rotatebox[origin=c]{-90}{AoLP}
    \\ \specialrule{0em}{1pt}{1pt}
    \textsc{Elephant} &
    \begin{subfigure}[c]{\imagesize\textwidth}
        \centering
        \includegraphics[width=\textwidth]{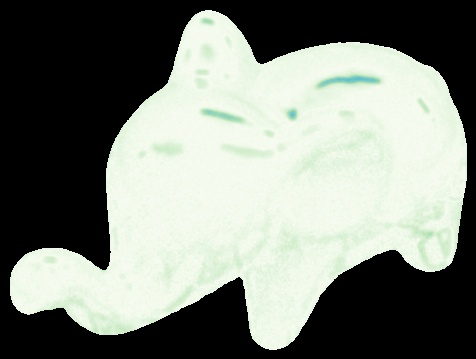}
    \end{subfigure} &
    \begin{subfigure}[c]{\imagesize\textwidth}
        \centering
        \includegraphics[width=\textwidth]{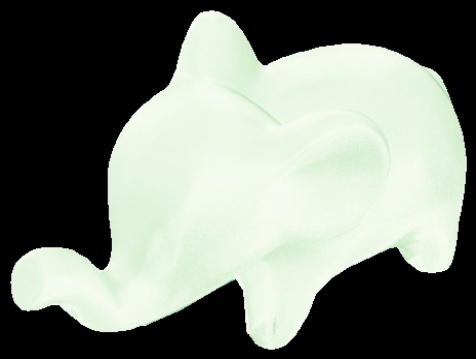}
    \end{subfigure} &
    \begin{subfigure}[c]{\imagesize\textwidth}
        \centering
        \includegraphics[width=\textwidth]{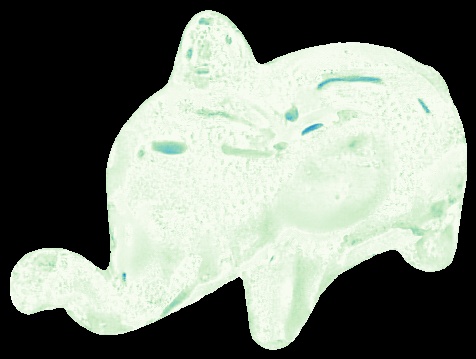}
    \end{subfigure} &
    \rotatebox[origin=c]{-90}{DoLP} 
    \\ \specialrule{0em}{1pt}{1pt}
    \begin{subfigure}[c]{\imagesize\textwidth}
        \centering
        \includegraphics[width=\textwidth]{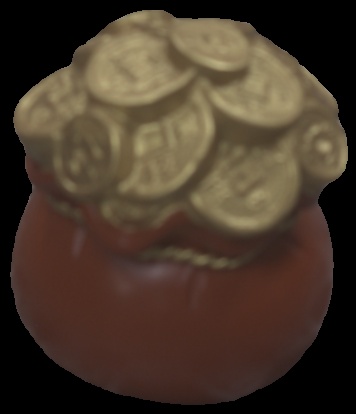}
    \end{subfigure} &
    \begin{subfigure}[c]{\imagesize\textwidth}
        \centering
        \includegraphics[width=\textwidth]{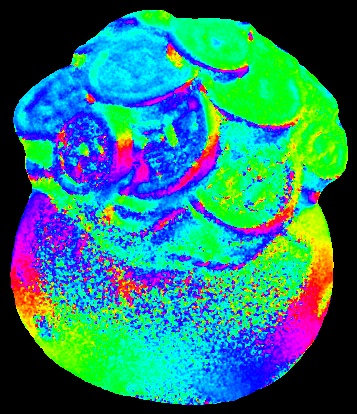}
    \end{subfigure} &
    \begin{subfigure}[c]{\imagesize\textwidth}
        \centering
        \includegraphics[width=\textwidth]{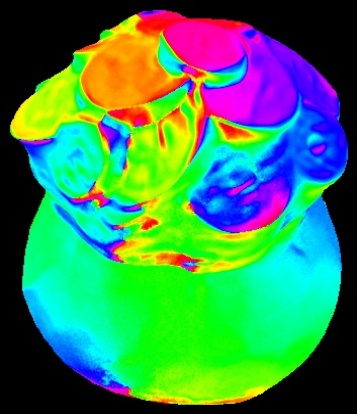}
    \end{subfigure} &
    \begin{subfigure}[c]{\imagesize\textwidth}
        \centering
        \includegraphics[width=\textwidth]{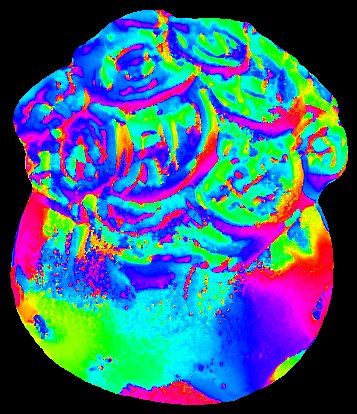}
    \end{subfigure} &
    \rotatebox[origin=c]{-90}{AoLP}
    \\ \specialrule{0em}{1pt}{1pt}
    \textsc{Money Jar} &
    \begin{subfigure}[c]{\imagesize\textwidth}
        \centering
        \includegraphics[width=\textwidth]{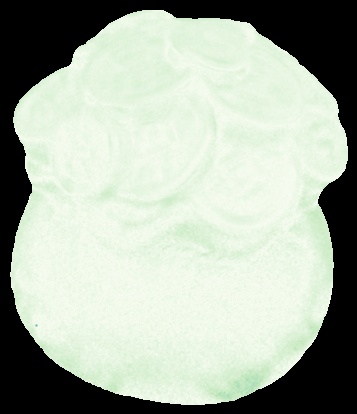}
    \end{subfigure} &
    \begin{subfigure}[c]{\imagesize\textwidth}
        \centering
        \includegraphics[width=\textwidth]{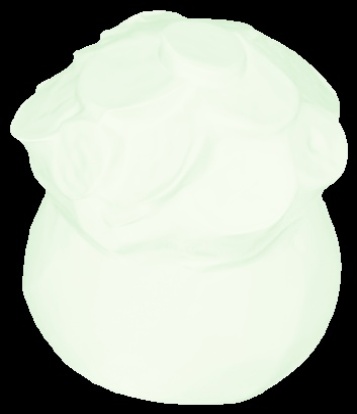}
    \end{subfigure} &
    \begin{subfigure}[c]{\imagesize\textwidth}
        \centering
        \includegraphics[width=\textwidth]{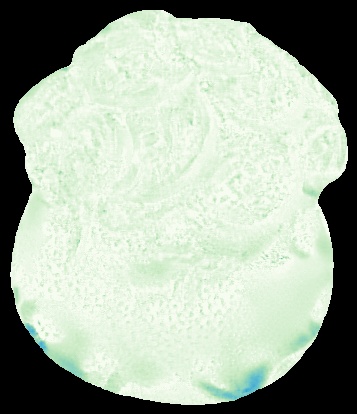}
    \end{subfigure} &
    \rotatebox[origin=c]{-90}{DoLP}
    \\
\end{tabular}
}
}
    \vspace{-3mm}
    \caption{Qualitative comparison with \mitsuba on selected views of {\sc Elephant} and {\sc Money Jar}.}
    \label{fig:eval_mitsuba_qual}
    \vspace{-5mm}
\end{figure}

\begin{figure*}[t]
    \centering
\renewcommand{\arraystretch}{0.2}
\newcommand{\imagesize}{0.14}
\newcommand{\halfsize}{0.07}
\resizebox{\linewidth}{!}{
\setlength{\tabcolsep}{1pt}{
\begin{tabular}{cc@{}c@{}c@{}c@{}c@{}c@{}c@{}c}
    & Diffuse & Dielectric & Metallic & Transparent & \makecell{Data from\\PANDORA} & \makecell{Data from\\NeRSP} & Scene 1 & Scene 2
    \\
    \raisebox{\halfsize\textwidth}{\rotatebox[origin=c]{90}{Input image}} 
    &
        
        \includegraphics[height=\imagesize\textwidth]{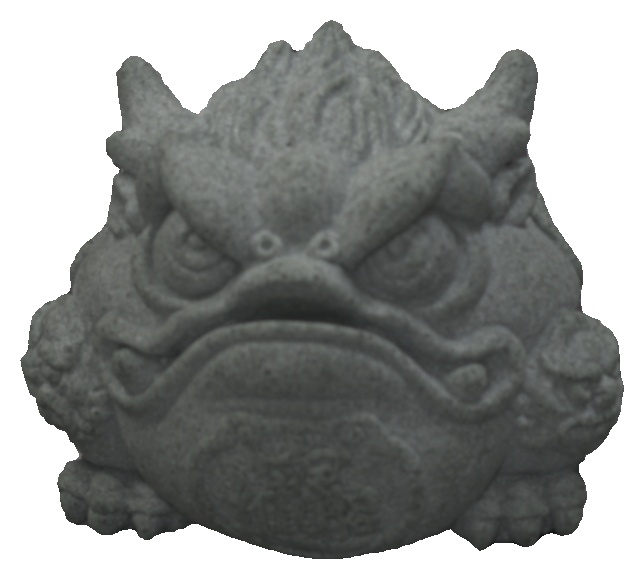}
    & 
        
        \includegraphics[height=\imagesize\textwidth]{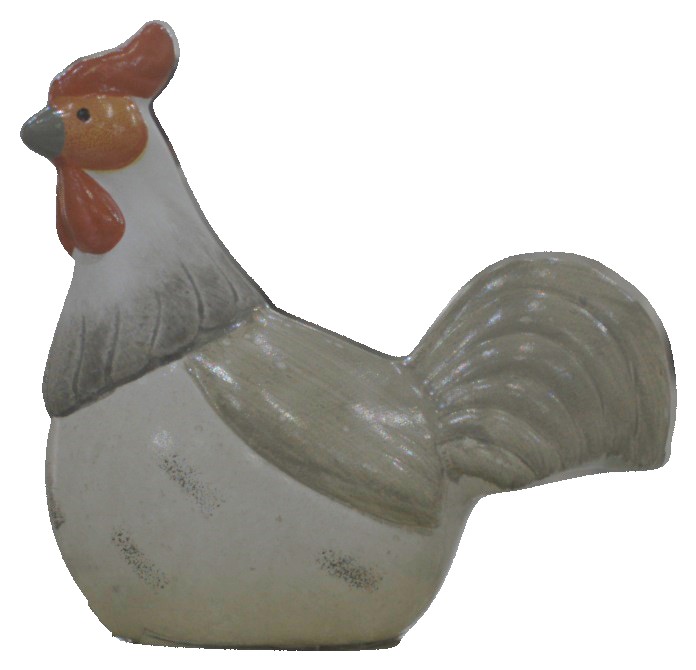}
    
    &
        
        \includegraphics[height=\imagesize\textwidth]{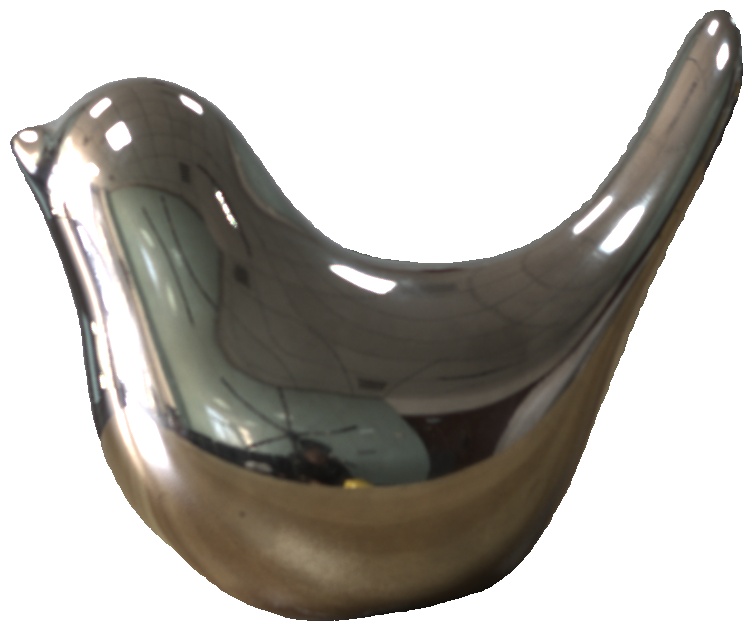}
    &
        
        \includegraphics[height=\imagesize\textwidth]{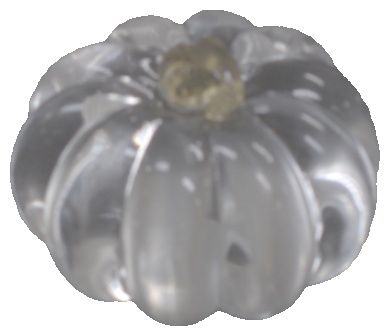}
    
    &
        
        \includegraphics[height=\imagesize\textwidth]{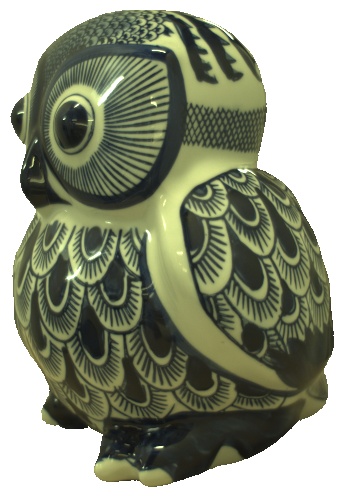}
    &
        
        \includegraphics[height=\imagesize\textwidth]{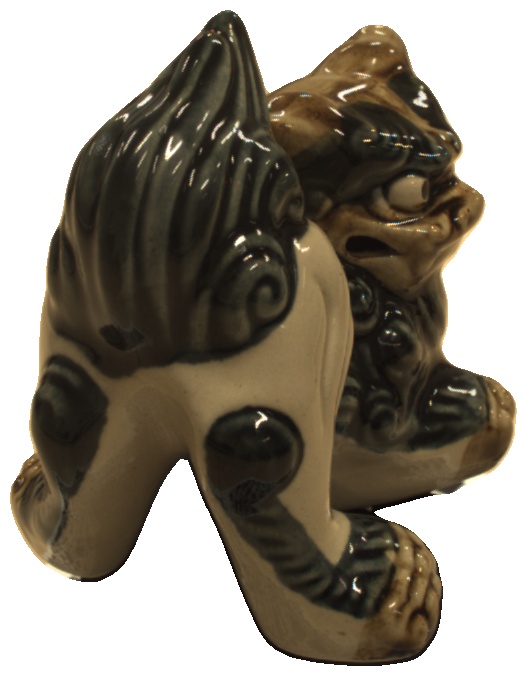}
    &
        
        \includegraphics[height=\imagesize\textwidth]{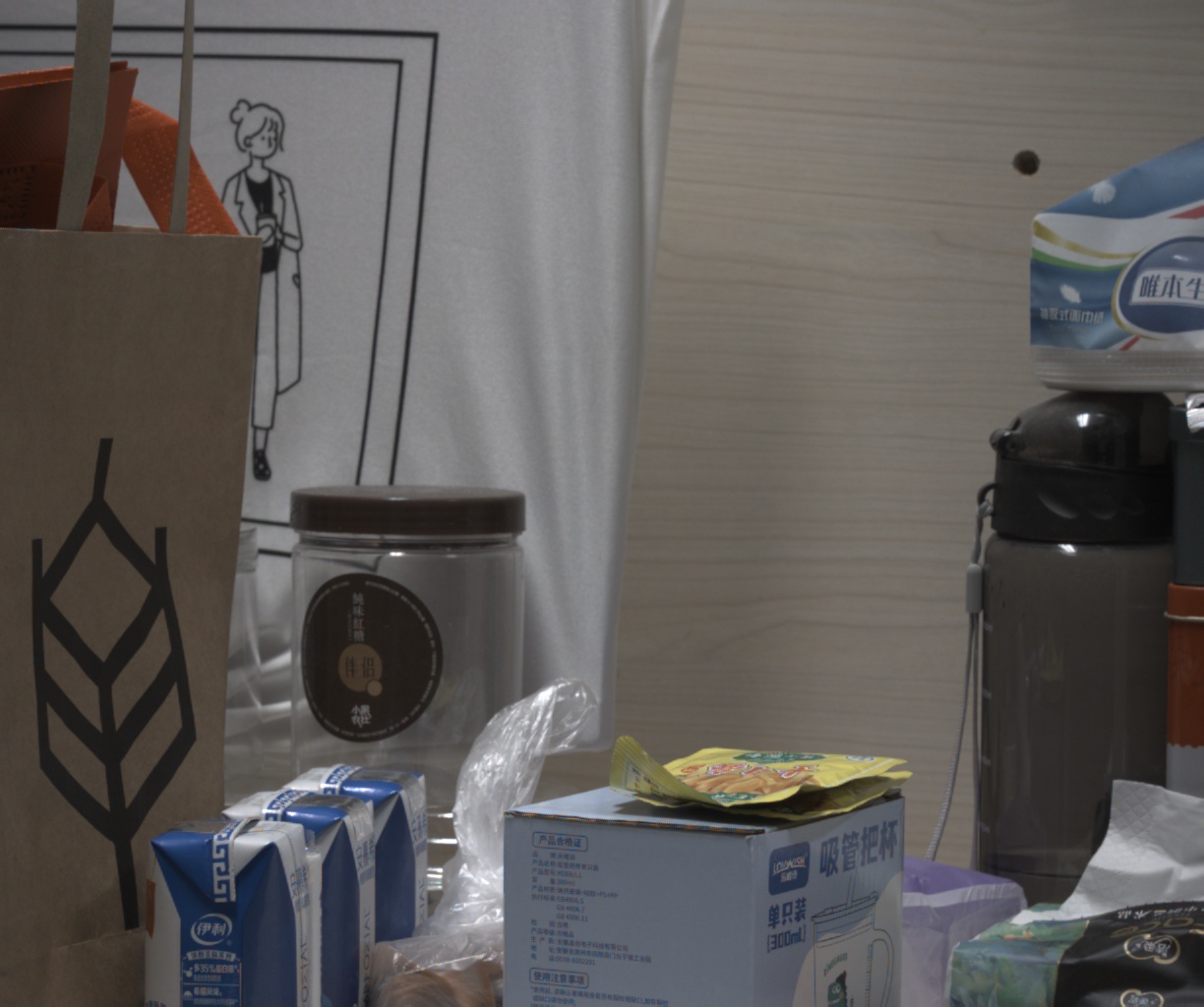}
    &
        
        \includegraphics[height=\imagesize\textwidth]{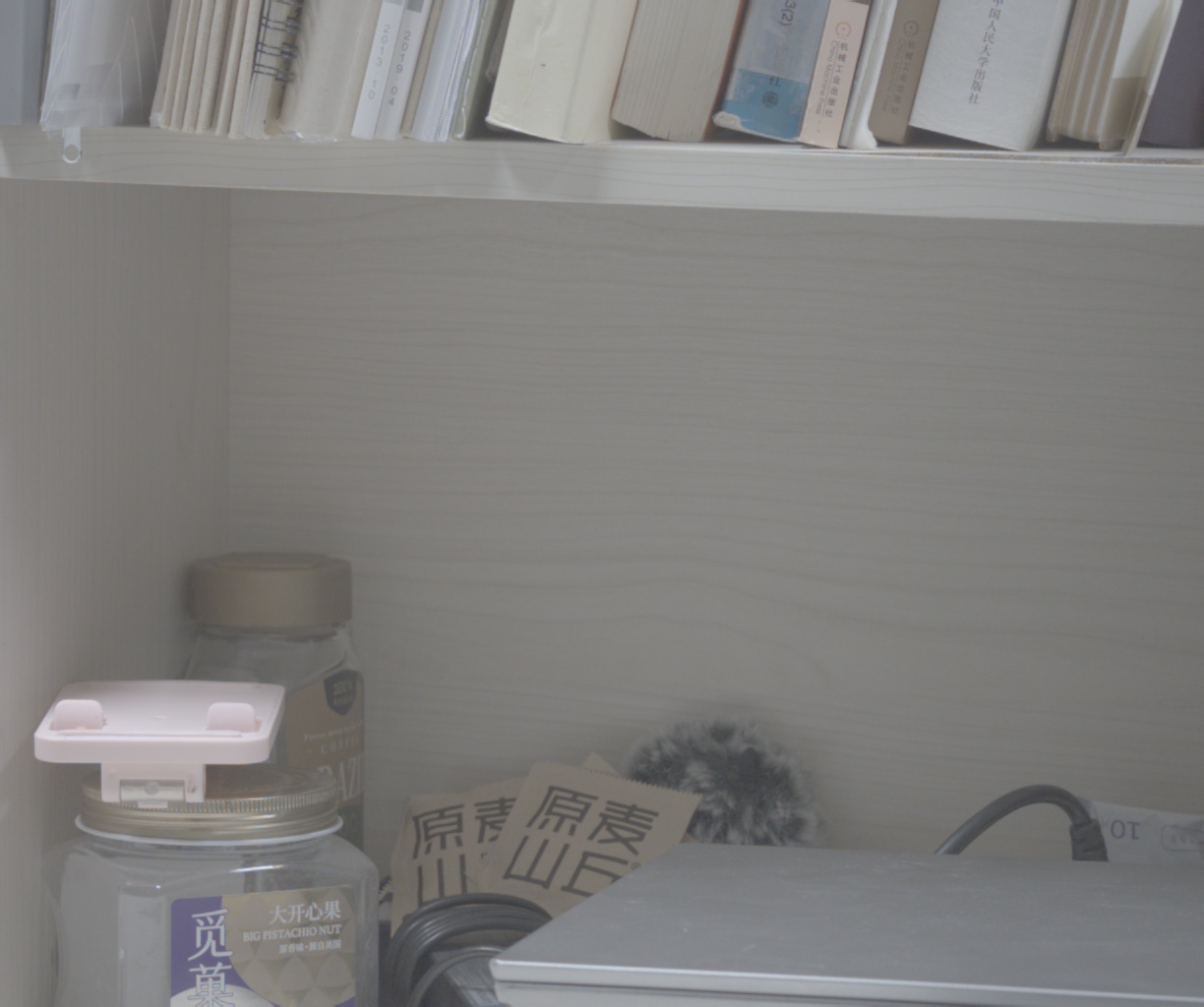}
    \\
    \raisebox{\halfsize\textwidth}{\rotatebox[origin=c]{90}{GT AoLP}} 
    & 
        \includegraphics[height=\imagesize\textwidth]{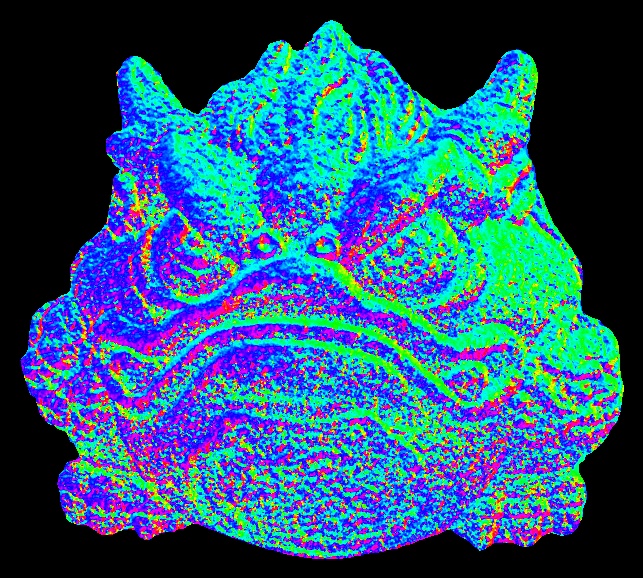}
    &
        \includegraphics[height=\imagesize\textwidth]{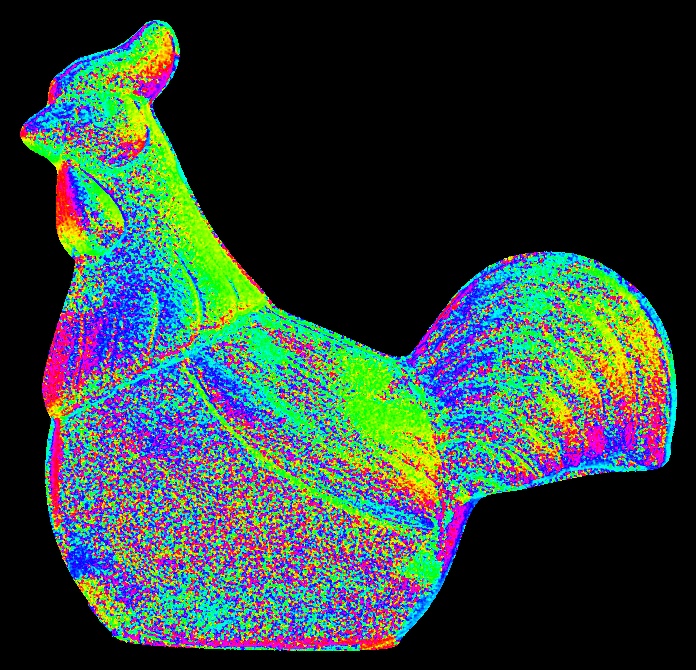}
    &
        \includegraphics[height=\imagesize\textwidth]{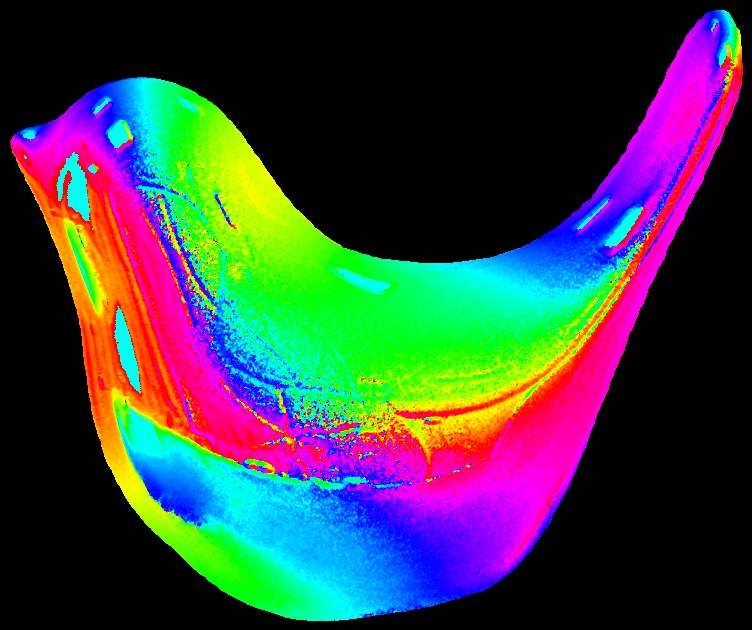}
    &
        
        \includegraphics[height=\imagesize\textwidth]{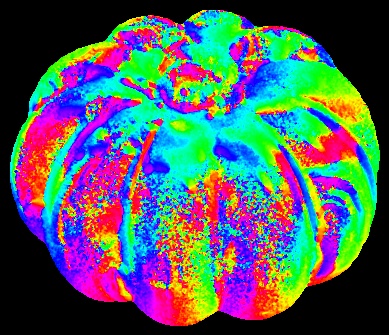}
    
    &
        
        \includegraphics[height=\imagesize\textwidth]{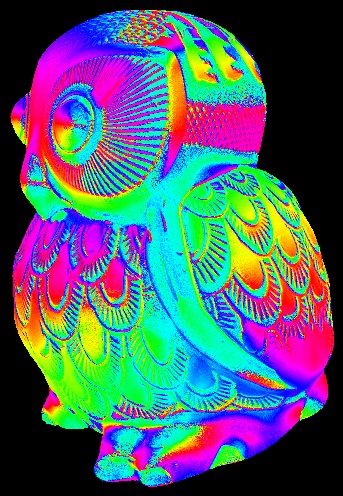}
    &
        
        \includegraphics[height=\imagesize\textwidth]{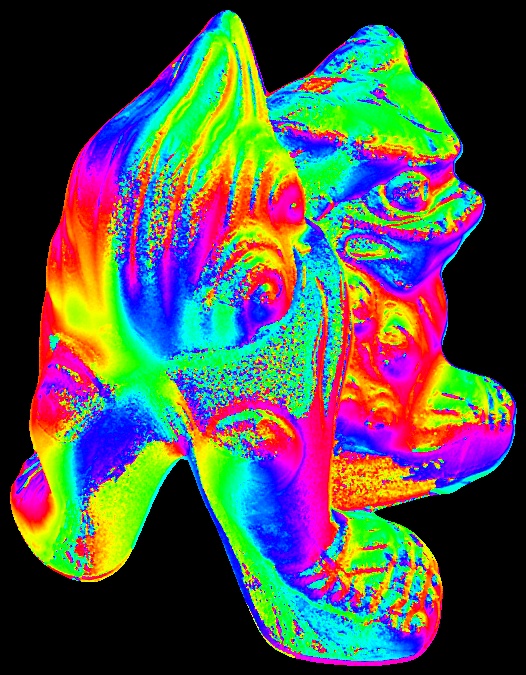}
    &
        
        \includegraphics[height=\imagesize\textwidth]{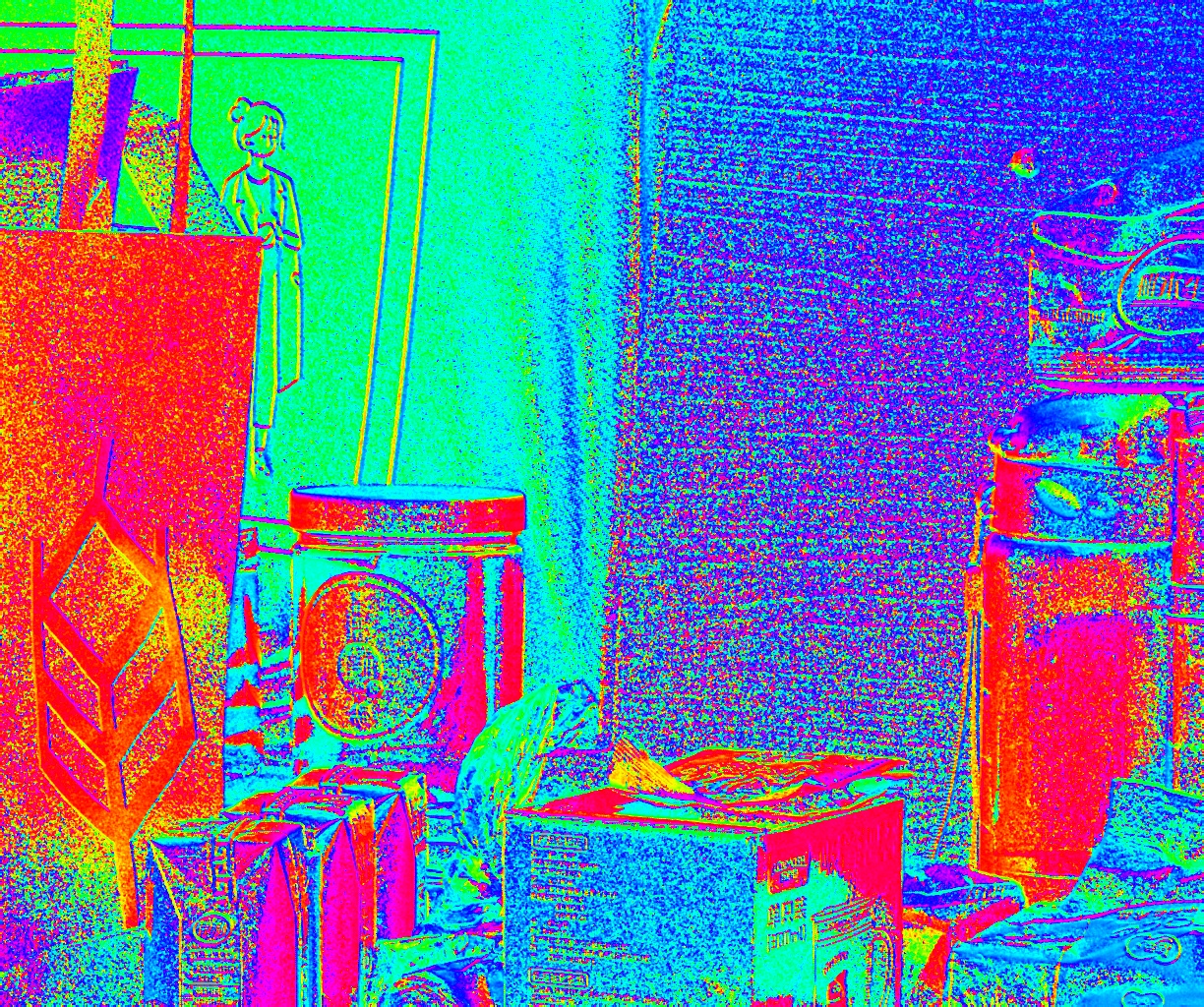}
    &
        
        \includegraphics[height=\imagesize\textwidth]{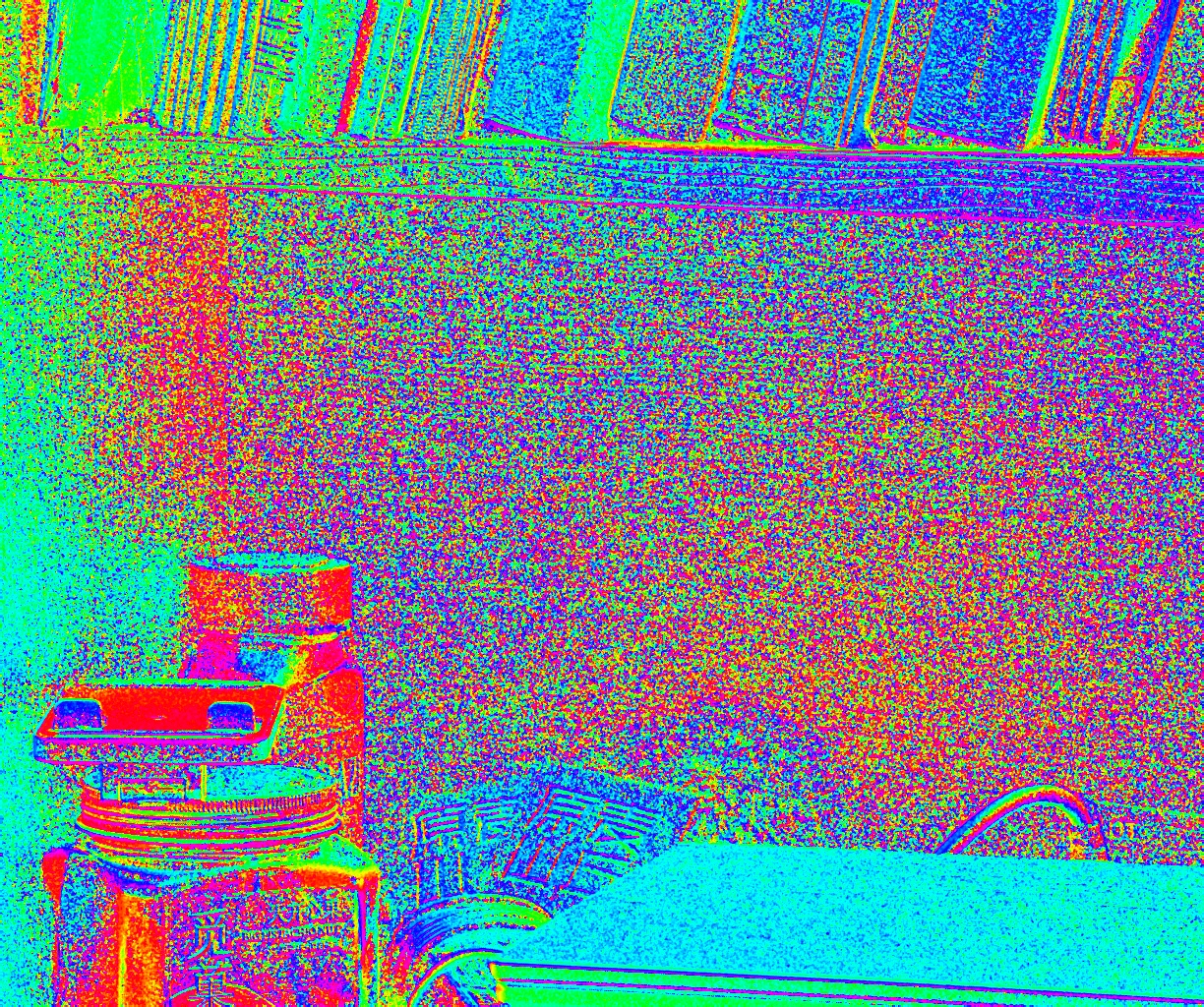}
    \\
    \raisebox{\halfsize\textwidth}{\rotatebox[origin=c]{90}{Gen. AoLP}} 
    
    & 
        \includegraphics[height=\imagesize\textwidth]{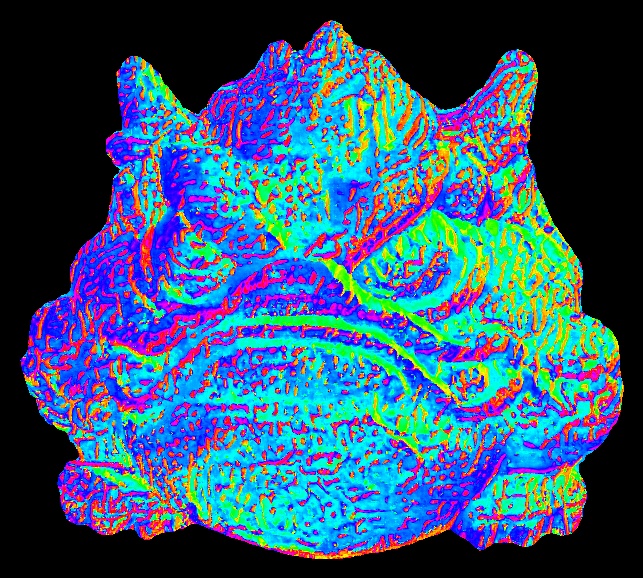}
    &
        
        \includegraphics[height=\imagesize\textwidth]{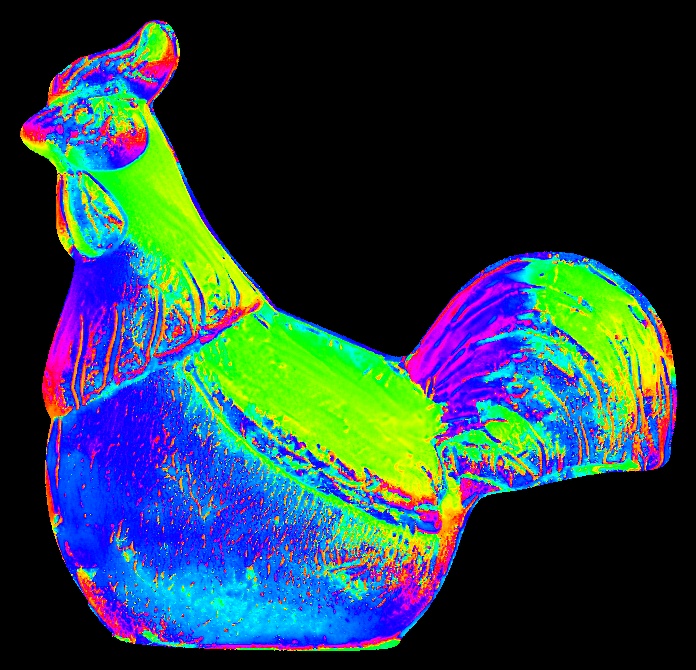}
    &
        
        \includegraphics[height=\imagesize\textwidth]{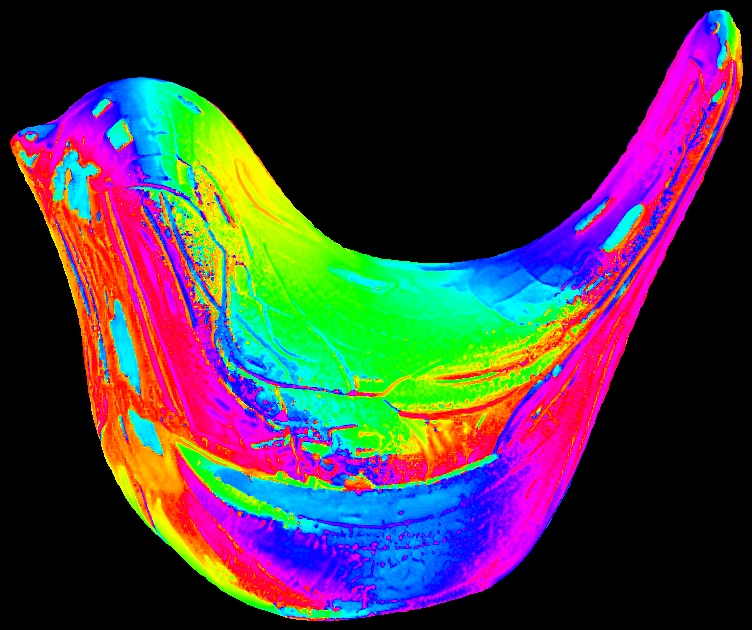}
    &
        
        \includegraphics[height=\imagesize\textwidth]{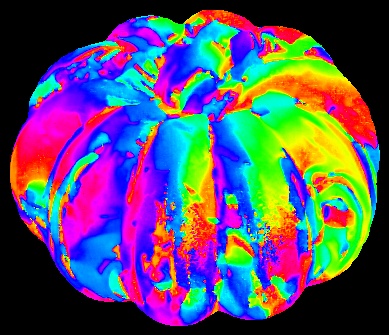}
    
    &
        
        \includegraphics[height=\imagesize\textwidth]{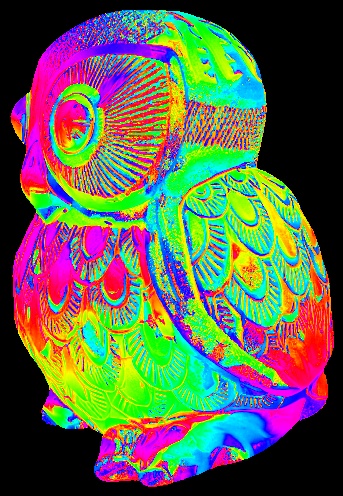}
    &
        
        \includegraphics[height=\imagesize\textwidth]{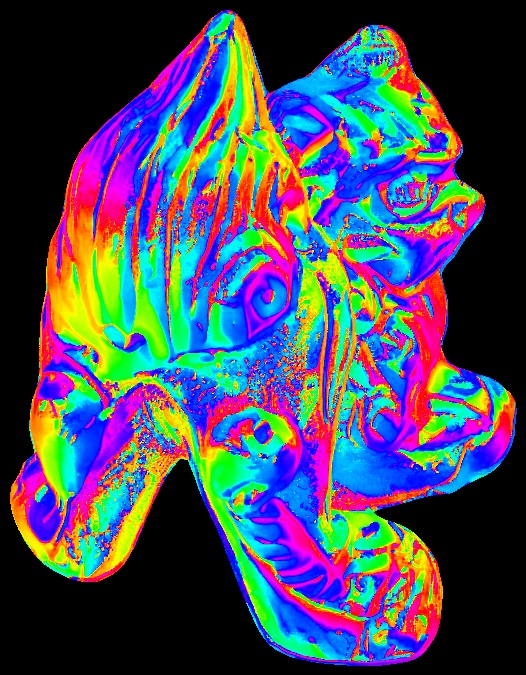}
    &
        
        \includegraphics[height=\imagesize\textwidth]{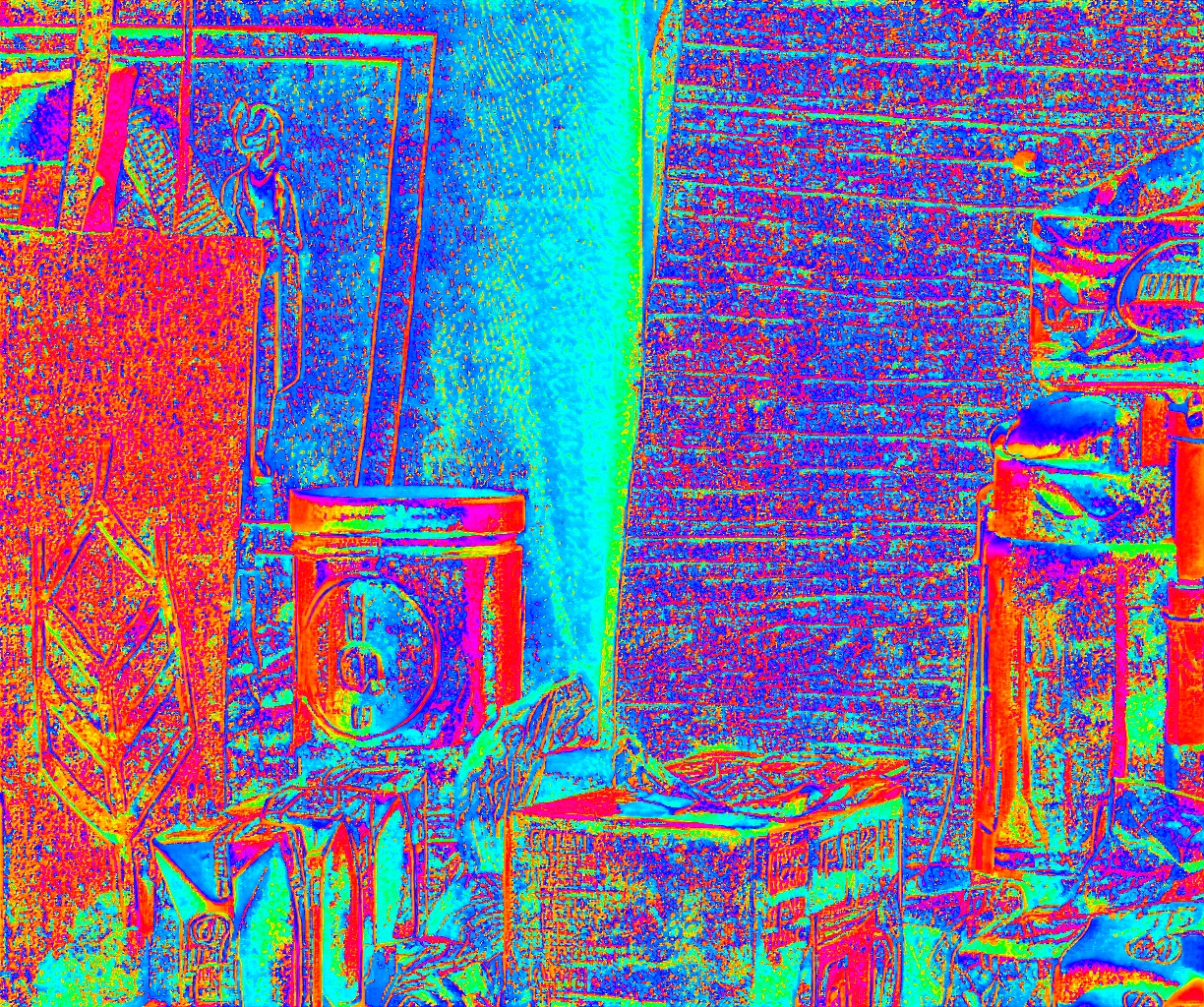}
    &
        
        \includegraphics[height=\imagesize\textwidth]{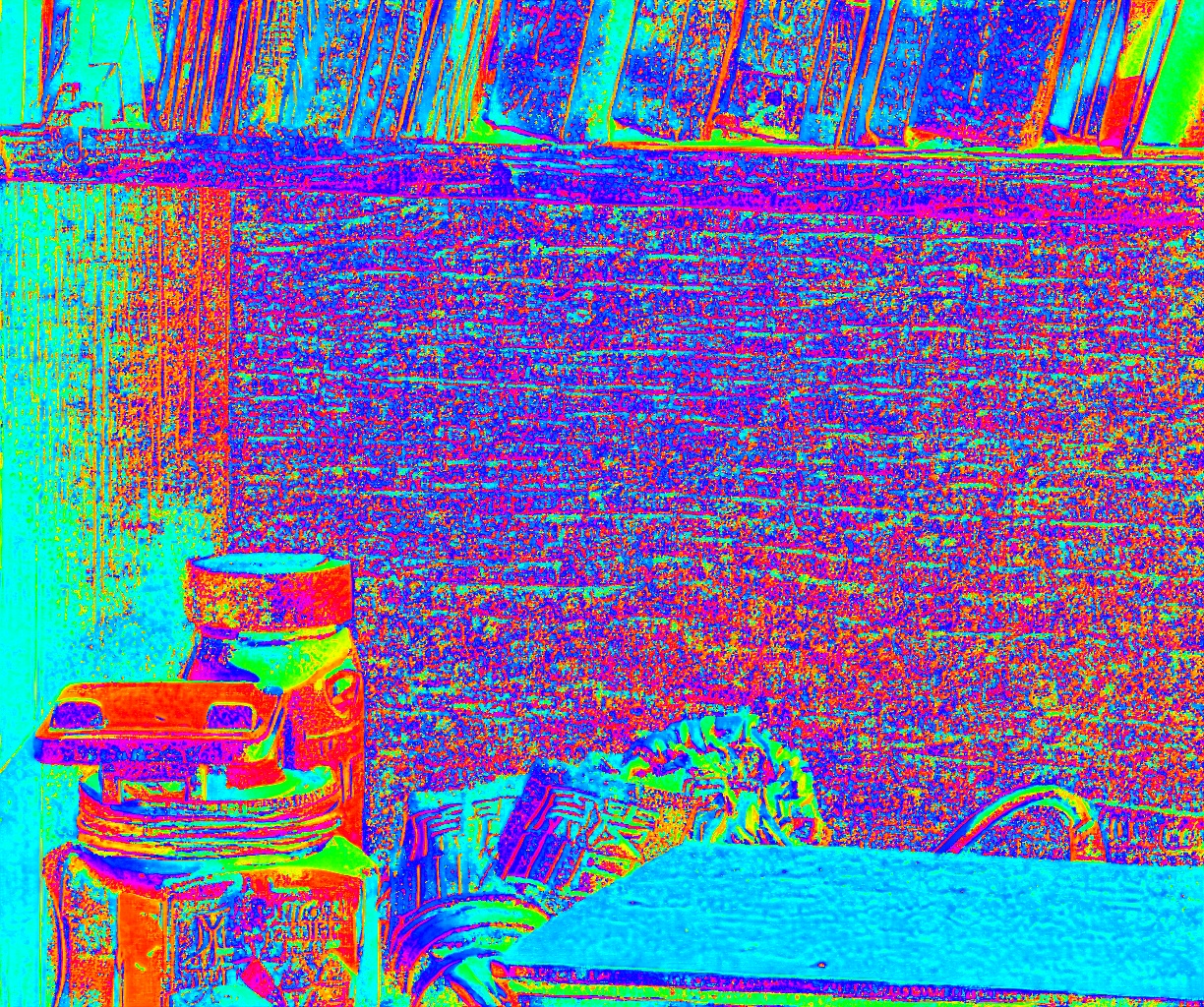}
    \\
    \raisebox{\halfsize\textwidth}{\rotatebox[origin=c]{90}{GT DoLP}} 
    & 
        \includegraphics[height=\imagesize\textwidth]{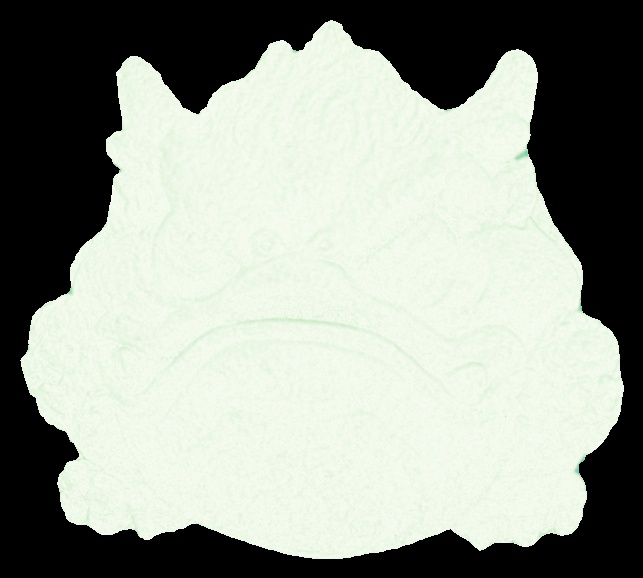}
    &
        
        \includegraphics[height=\imagesize\textwidth]{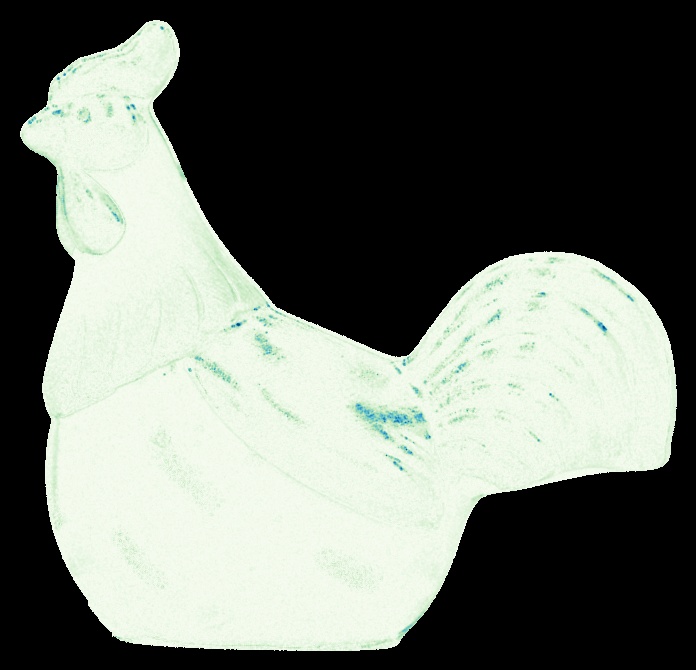}
    &

        \includegraphics[height=\imagesize\textwidth]{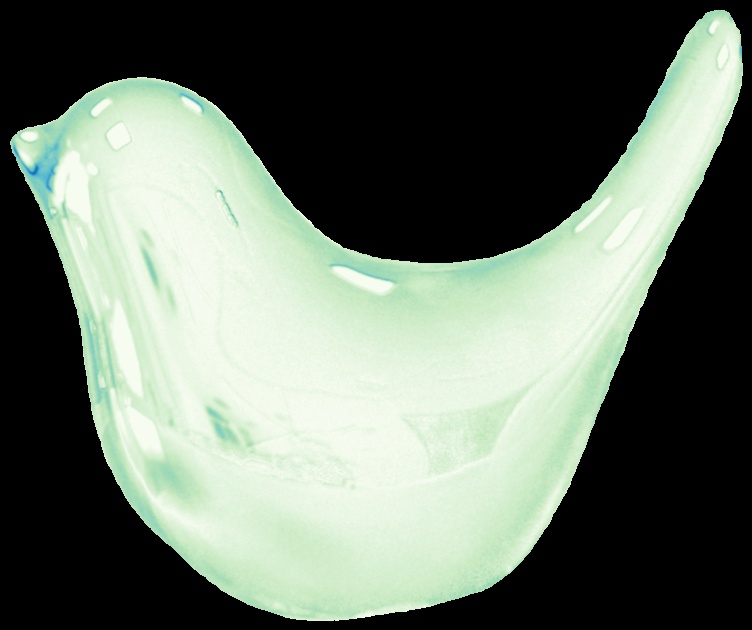}
    
    &
        
        \includegraphics[height=\imagesize\textwidth]{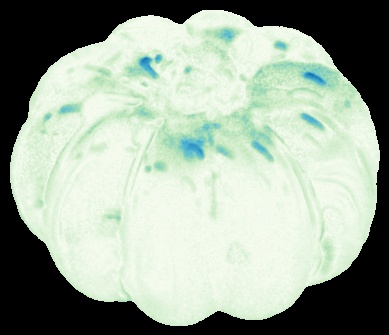}
    &
        
        \includegraphics[height=\imagesize\textwidth]{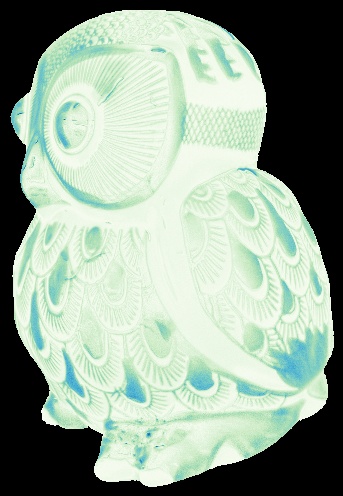}
    &
        
        \includegraphics[height=\imagesize\textwidth]{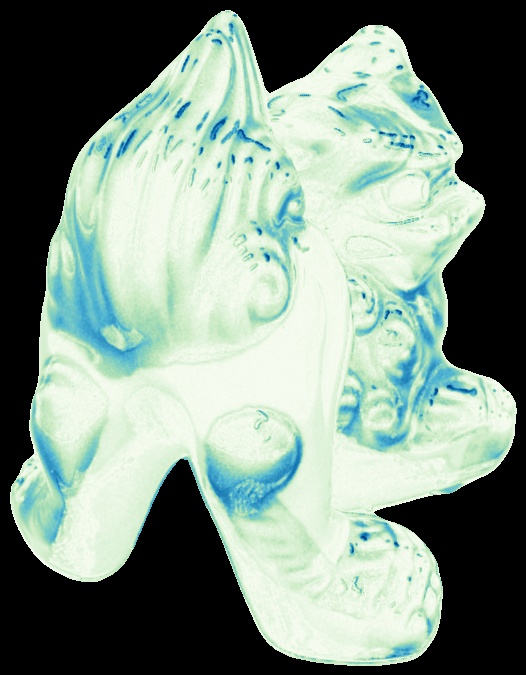}
    
    &
        
        \includegraphics[height=\imagesize\textwidth]{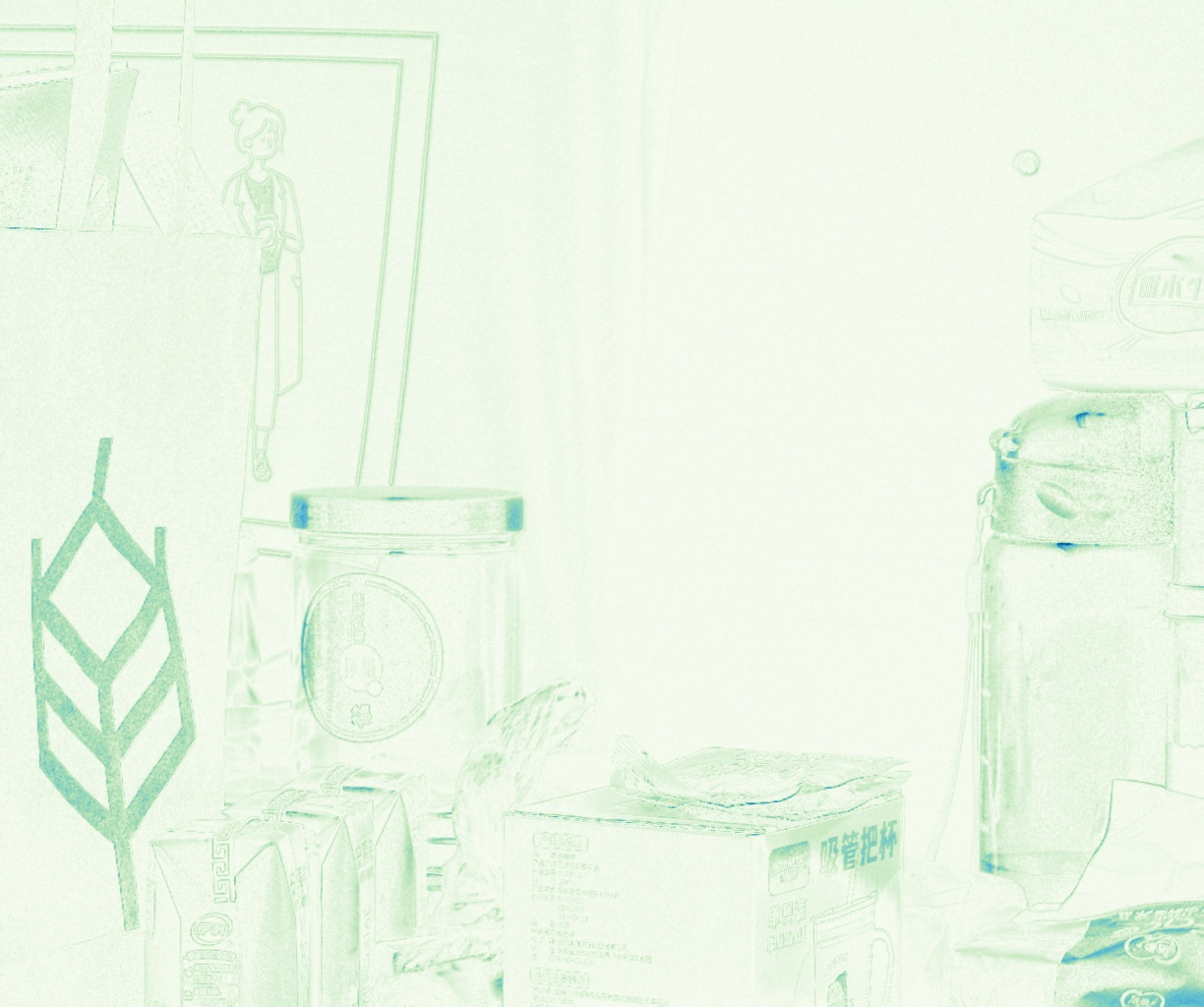}
    &
        
        \includegraphics[height=\imagesize\textwidth]{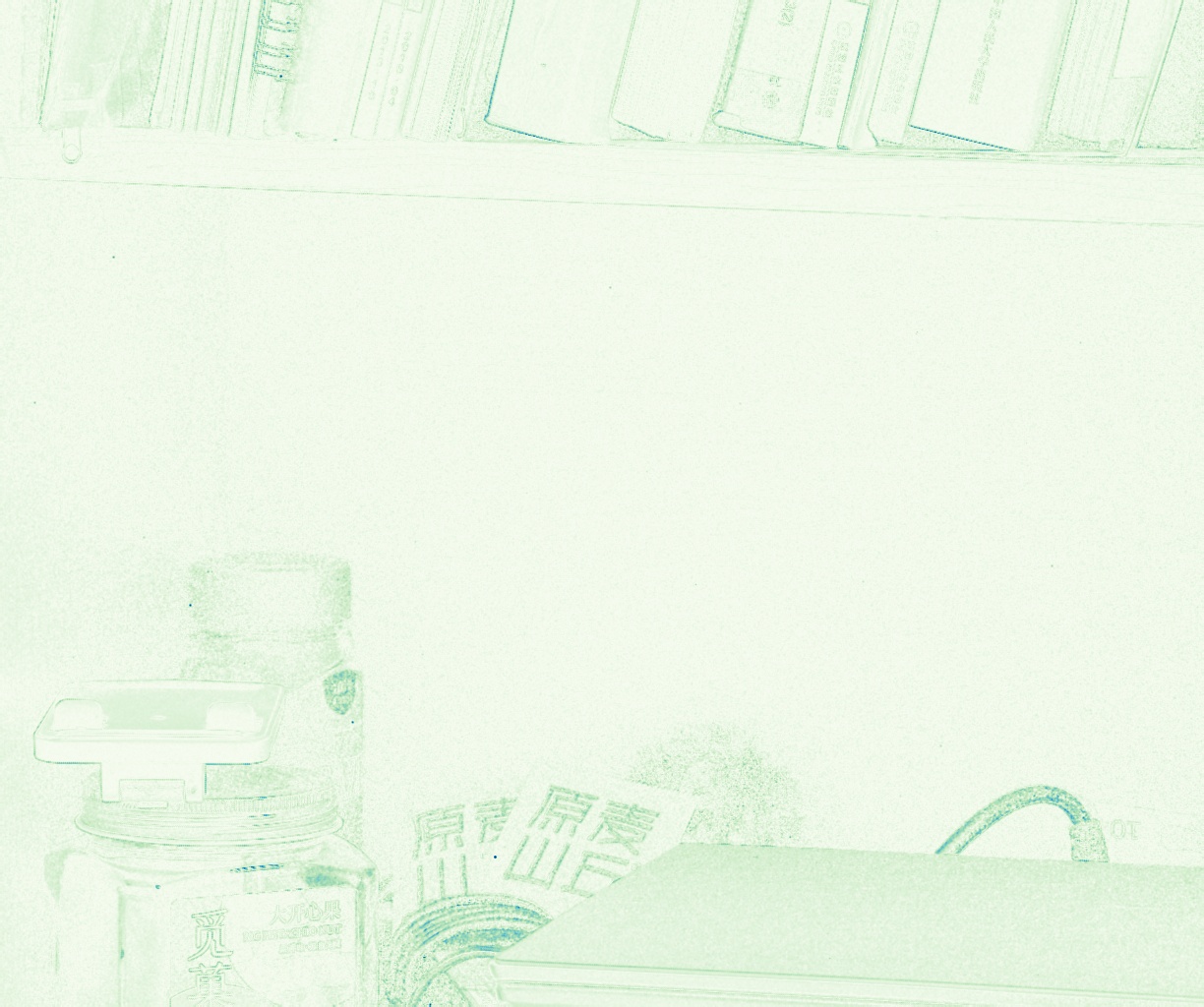}
    \\
    \raisebox{\halfsize\textwidth}{\rotatebox[origin=c]{90}{Gen. DoLP}} 
    & 
        \includegraphics[height=\imagesize\textwidth]{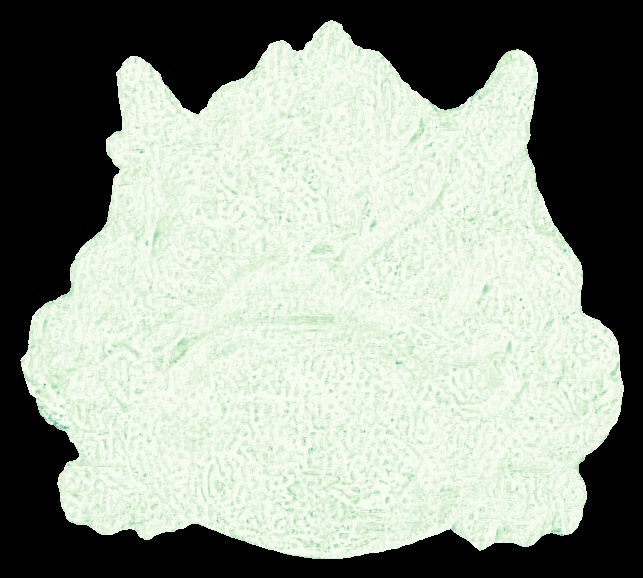}
    &
        \includegraphics[height=\imagesize\textwidth]{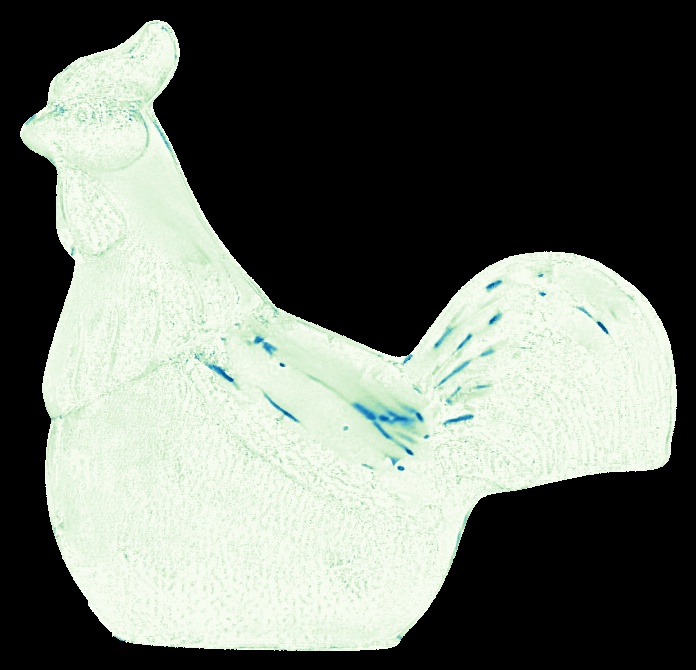}
    &
        \includegraphics[height=\imagesize\textwidth]{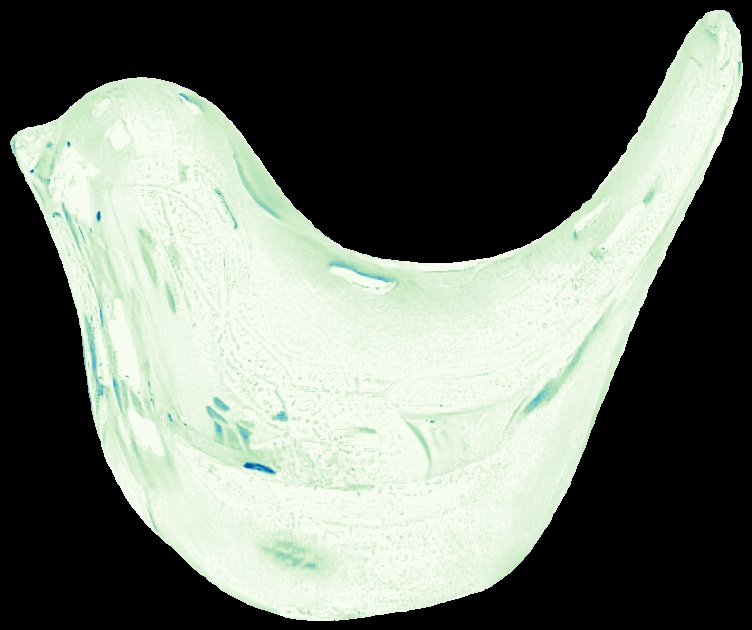}
    &
        
        \includegraphics[height=\imagesize\textwidth]{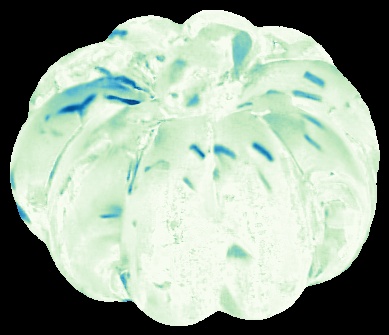}
    
    &
        
        \includegraphics[height=\imagesize\textwidth]{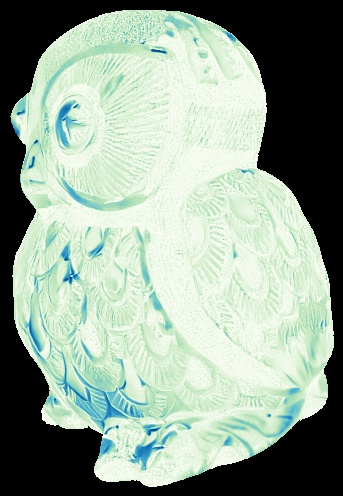}
    &
        
        \includegraphics[height=\imagesize\textwidth]{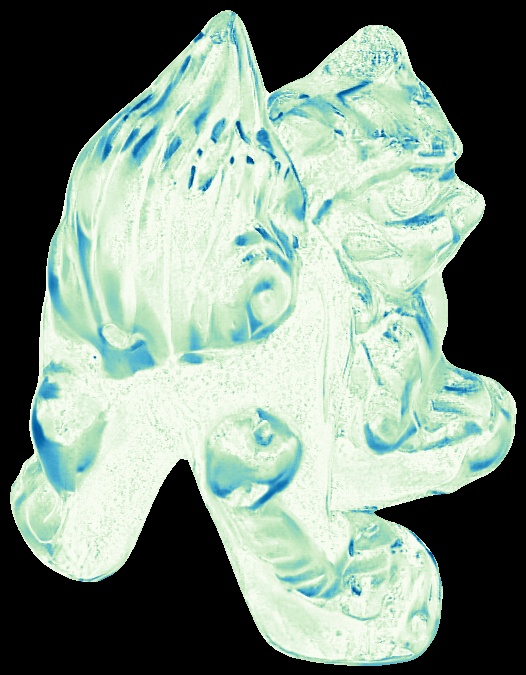}
    &
        
        \includegraphics[height=\imagesize\textwidth]{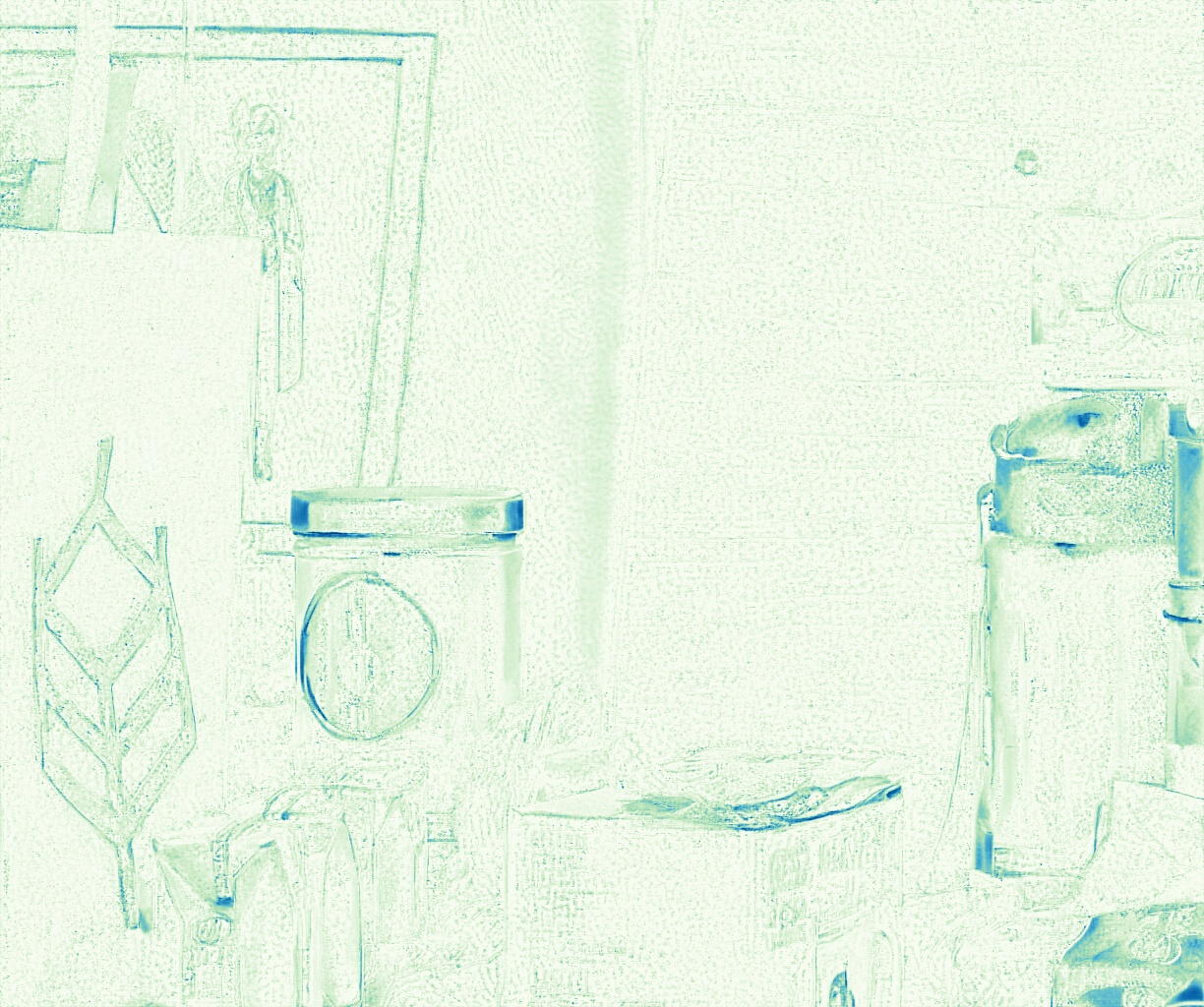}
    &
        
        \includegraphics[height=\imagesize\textwidth]{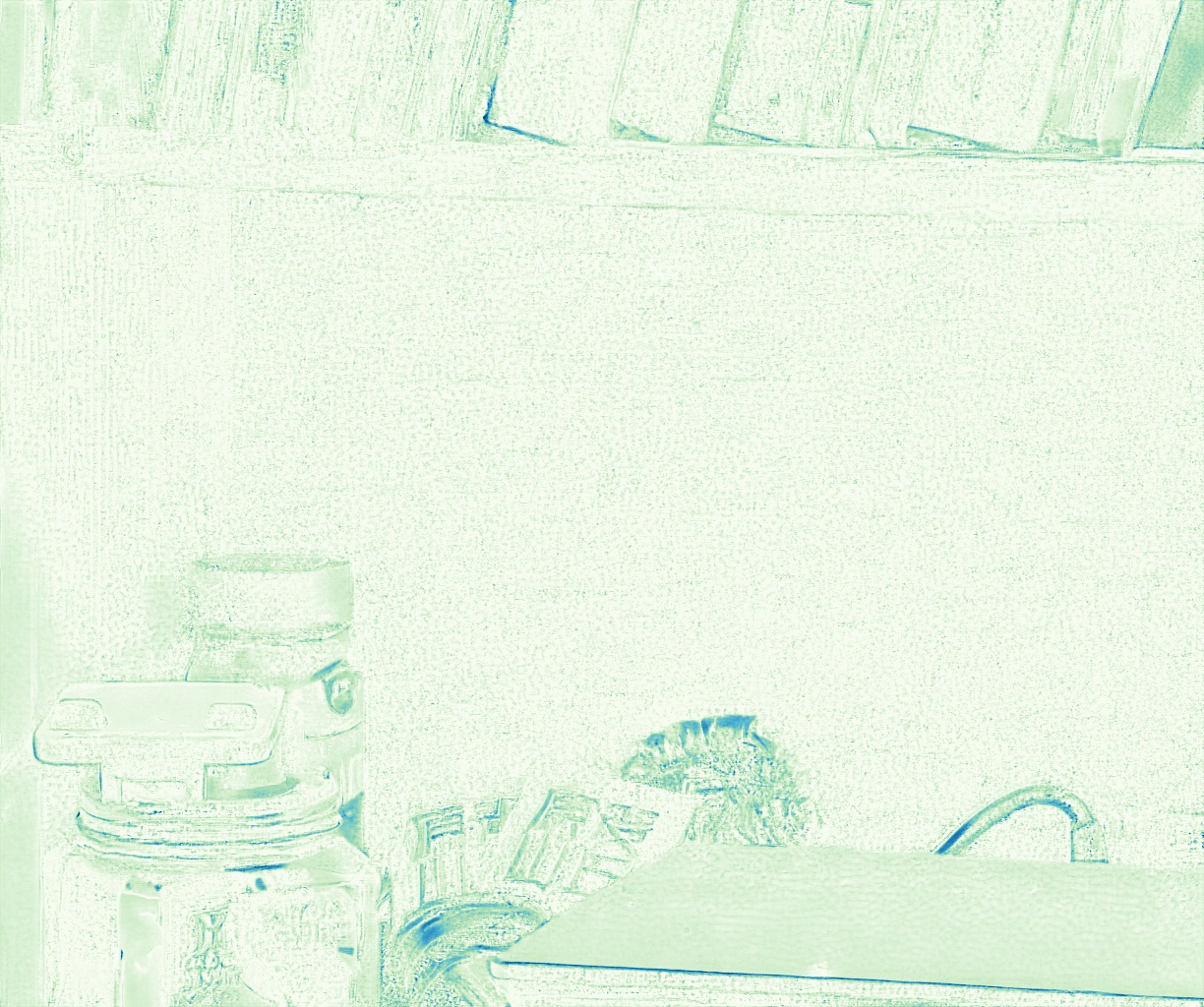}
    \\
\end{tabular}
}
}

    \vspace{-3mm}
    \caption{Qualitative evaluation on causal captured images with diverse shape and reflectance. The method works robustly on different materials (Columns 1-4), public polarization test datasets (Columns 5-6), and scene-level polarization synthesis (Columns 7-8).}
    \label{fig:eval_causal_qual}
\end{figure*}
\paragraph{Comparison with Mitsuba}
We compare the quality of polarization images synthesized from our generative model \pa and physics-based renderer \mitsuba. For polarization simulation, Mitsuba requires environment maps, surface meshes, BRDF, and physically based rendering (PBR) parameters such as roughness and albedo. \Fref{fig:pipeline_cmp_mit_pa} illustrates the polarization generation pipelines of Mitsuba and \pa. To reproduce polarization images using Mitsuba, we capture multi-view polarization images of a dielectric object {\sc Elephant} and a diffuse object {\sc Money Jar}, each set contains approximately 26 views with calibrated camera poses with Metashape~\cite{metashape}. Applying the state-of-the-art neural inverse rendering technique \nero on unpolarized components of captured images, we obtain the corresponding shapes, environment maps, and the PBR parameter maps. Given these inputs, we can render polarization images using \mitsuba for each view. On the other hand, \pa only takes the unpolarized component of each polarization image as input and generates the polarized ones.
\Fref{fig:eval_mitsuba_qual} compares \pa against Mitsuba on generated AoLP and DoLP maps. Despite taking only a single RGB image as input, \pa produces AoLP and DoLP much closer to the ground truth, achieving consistent performance across both specular ({\sc Elephant}) and diffuse ({\sc Money Jar}) surface types. 
Quantitative evaluation is infeasible in this experiment due to misalignment between the Mitsuba-rendered images and the GT ones, which is attributed to camera calibration errors and inherent inaccuracies in \nero.

\vspace{-5.4mm}
\paragraph{Comparison with real polarization images}
\Fref{fig:eval_causal_qual} displays the \pa generated polarization properties compared to the real-captured ones. We show that \pa can handle a wide range of materials, including diffuse, dielectric, metallic, and even transparent surfaces. Columns 1 to 4 show the results on our test set.
In the $5$-th and $6$-th columns, we test \pa on public real-world polarization image datasets released by \nersp and \pandora. Though the two datasets may have a distribution gap compared to our captured dataset for training, \pa still demonstrates its generalization ability. As shown in the last two columns, \pa is capable of generating realistic scene-level polarization properties, which are challenging for parametric polarization image simulators like \mitsuba due to the scarcity of scene-level 3D assets with consistent geometry and PBR materials. Our method can easily simulate a polarization image with a causal capture of the target scene. 

\begin{figure*}

    \centering
    \begin{overpic}[width=0.95\linewidth]{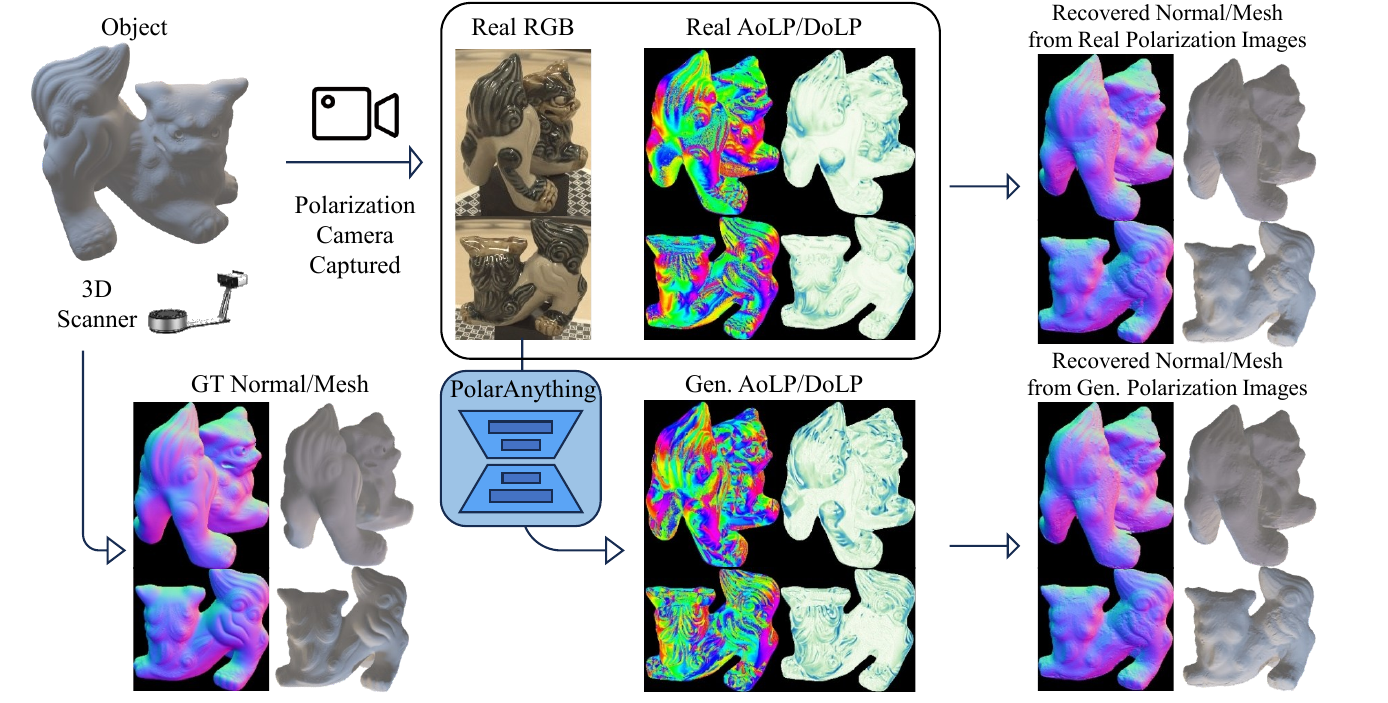}
    \put(68.9,12.7){\color{black}{\fontsize{8pt}{1pt}\selectfont PISR}}
    \put(68.9,38.5){\color{black}{\fontsize{8pt}{1pt}\selectfont PISR}}
    \end{overpic}
    \small{
    \begin{tabular}{llcccc}
\toprule
    Object & Inputs for \pisr  & MAngE~(Normal) & CD~(Mesh) & MAngE~(AoLP) & MAbsE~(DoLP) \\
    \midrule
    \multirow{2}{*}{\sc Shisa} & Real Pol. Images & 15.45 & 0.6765 & N/A & N/A \\
    & Gen. Pol. Images by \pa & 15.17 & 0.6564 & 33.68 & 0.1563 \\
\bottomrule
\end{tabular}
    }
    \vspace{-2mm}
    \caption{Evaluating \pa on multiview shape from polarization based on \pisr. 
    }
    \label{fig:eval_pisr}
\end{figure*}

\subsection{Application: Boosting Shape from Polarization}
Shape from polarization (SfP) reconstructs 3D surfaces from single- or multi-view polarization images by leveraging the geometric cues encoded in AoLP and DoLP. 
With \pa, we show 1) existing multi-view SfP methods can be directly applied to RGB images by first converting them into synthesized polarization images, achieving shape reconstruction results comparable to those using real polarization data; and 2) by integrating \pa with Stanford-ORB~\cite{kuang2023stanfordorb} containing RGB-normal pairs, we introduce \textbf{PolarStanford-ORB}, a dataset comprising synthesized polarization images and their corresponding surface normals. This dataset benefits learning-based SfP methods by expanding the training set while mitigating the need for costly polarization image acquisition and ground-truth surface normal collection.

\begin{figure*}[t]
    \newcommand{\imagesize}{0.125}
\resizebox{\linewidth}{!}{
\setlength{\tabcolsep}{1pt}{
\begin{tabular}{cc@{}cc@{}cc@{}cc}
    Input & \multicolumn{2}{c}{\deepsfp} & \multicolumn{2}{c}{\makecell{DeepSfP+MSO}} & \multicolumn{2}{c}{\makecell{DeepSfP+PSO}} & GT
    \\
    \includegraphics[width=\imagesize\textwidth]{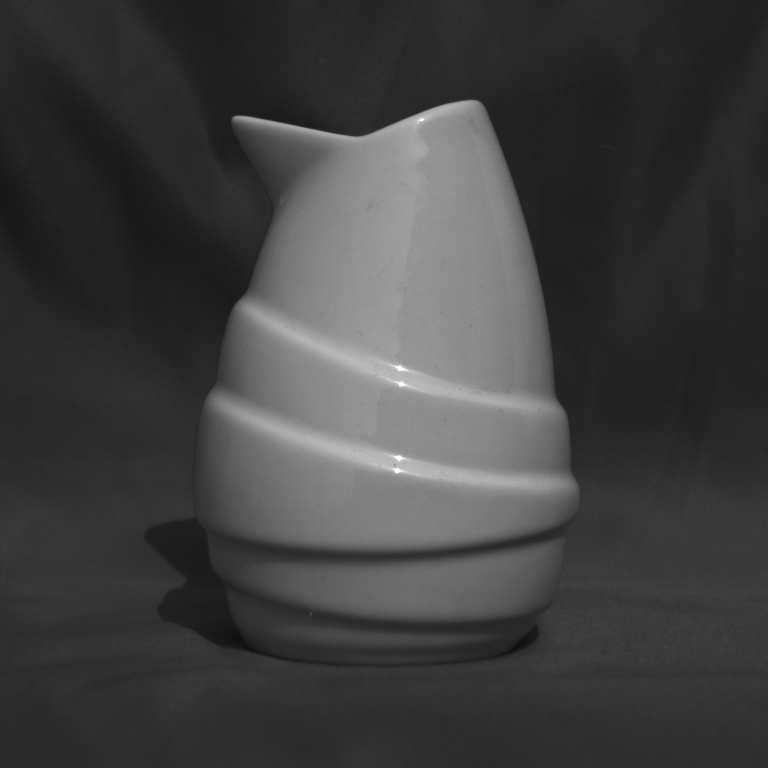}
    &
    \begin{overpic}[percent,width=\imagesize\textwidth]{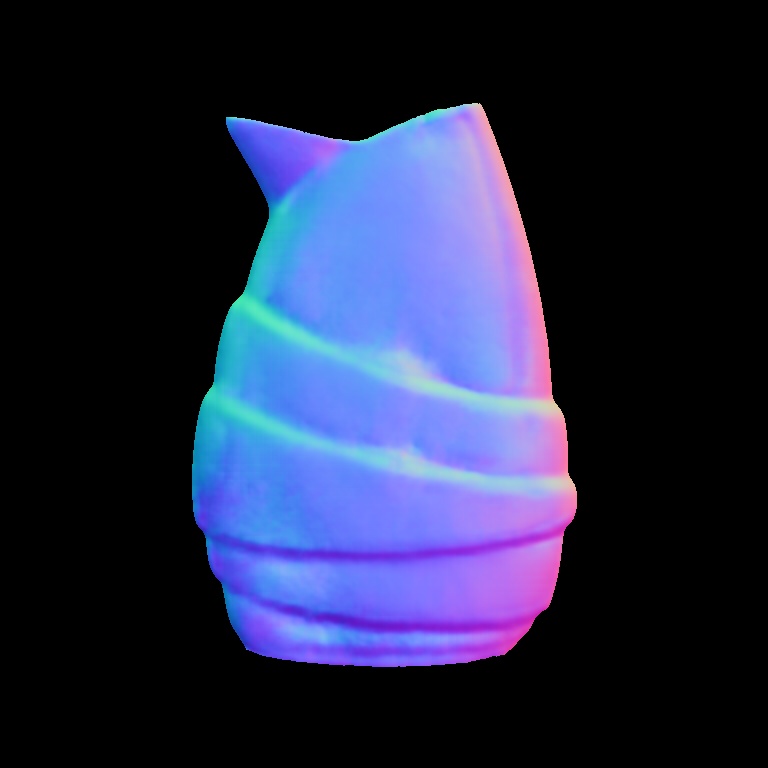}
    \put(80,4){\color{white}\scriptsize{14.23}}
    \end{overpic}
    &
    \includegraphics[width=\imagesize\textwidth]{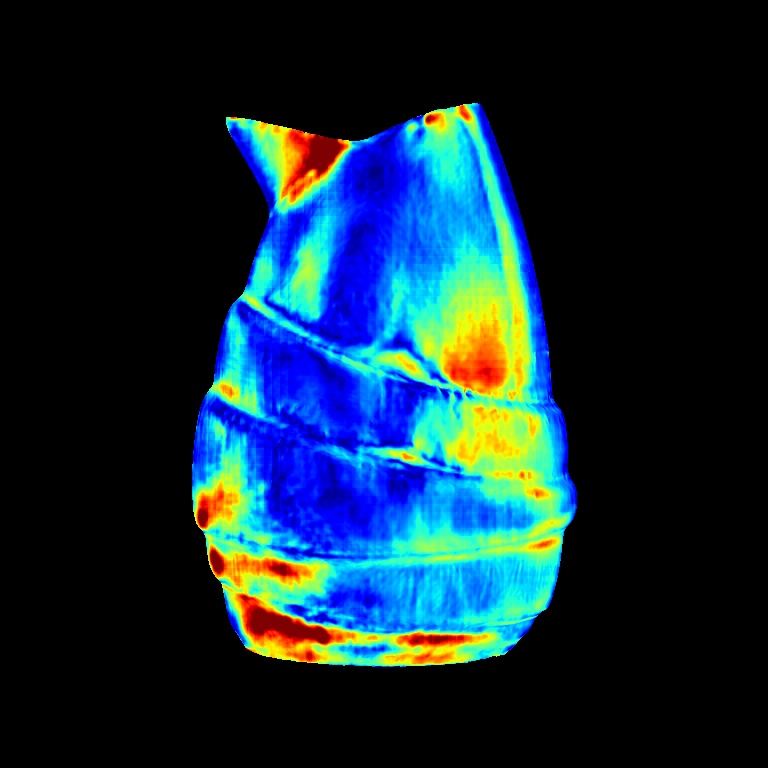}
    &
    \begin{overpic}[percent,width=\imagesize\textwidth]{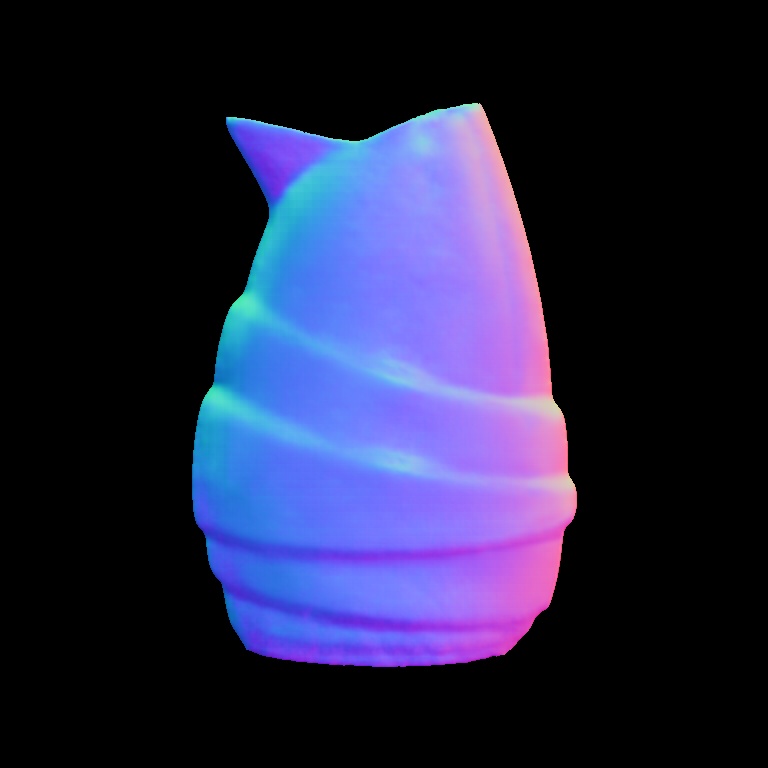}
    \put(80,4){\color{white}\scriptsize{11.94}}
    \end{overpic}
    &
    \includegraphics[width=\imagesize\textwidth]{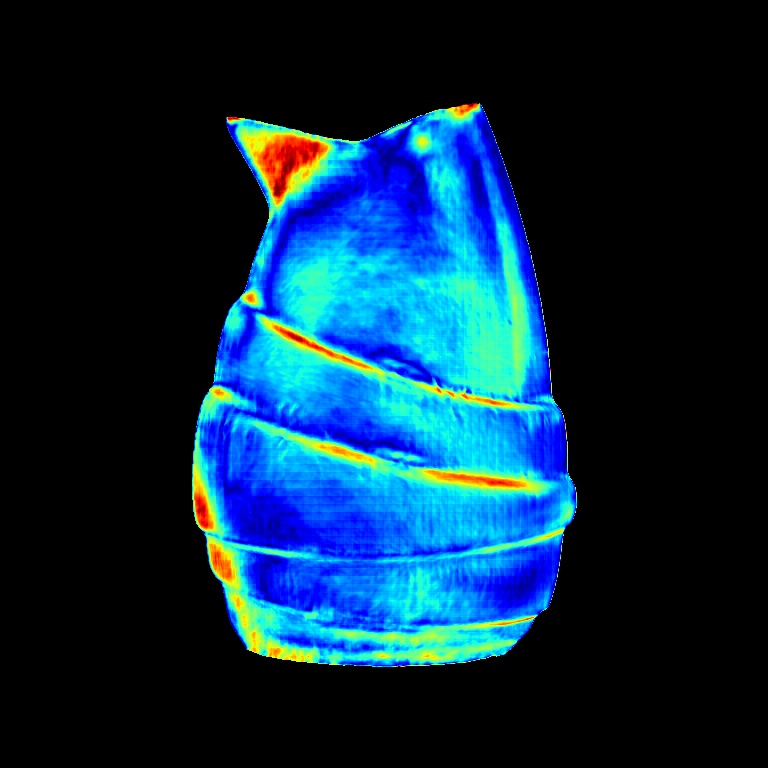}
    &
    \begin{overpic}[percent,width=\imagesize\textwidth]{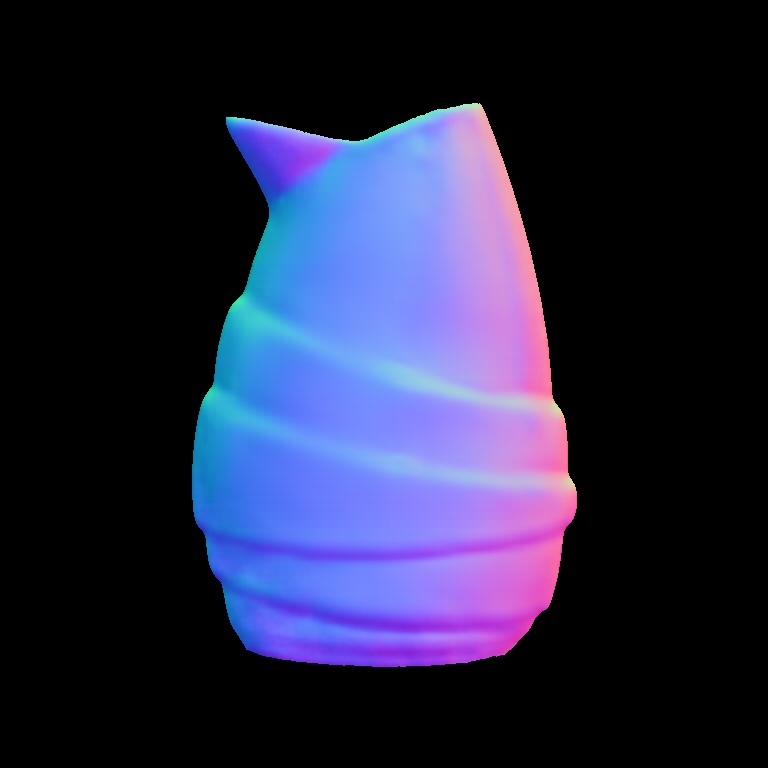}
    \put(80,4){\color{white}\scriptsize{9.05}}
    \end{overpic}
    &
    \includegraphics[width=\imagesize\textwidth]{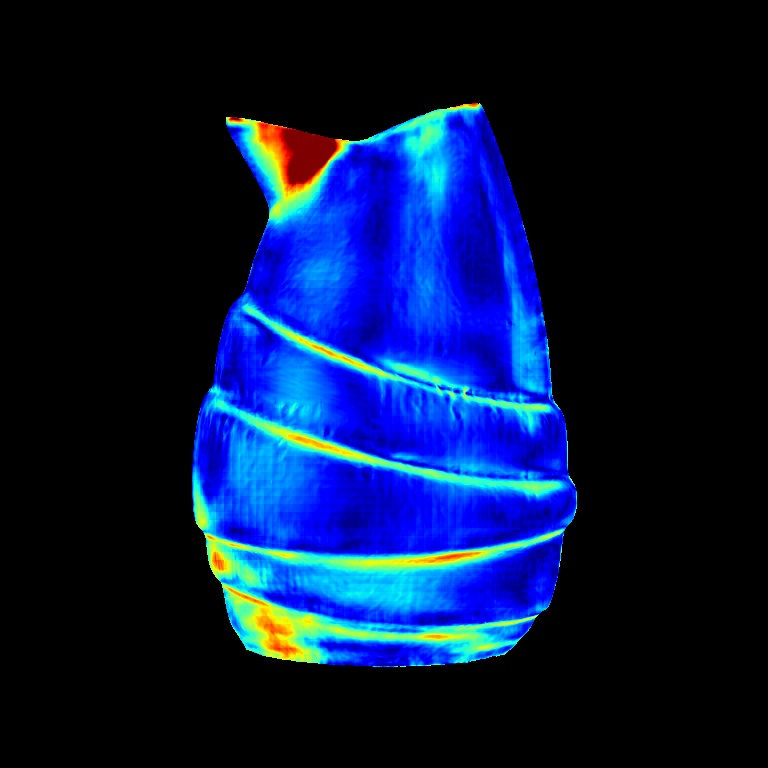}
    &
    \includegraphics[width=\imagesize\textwidth]{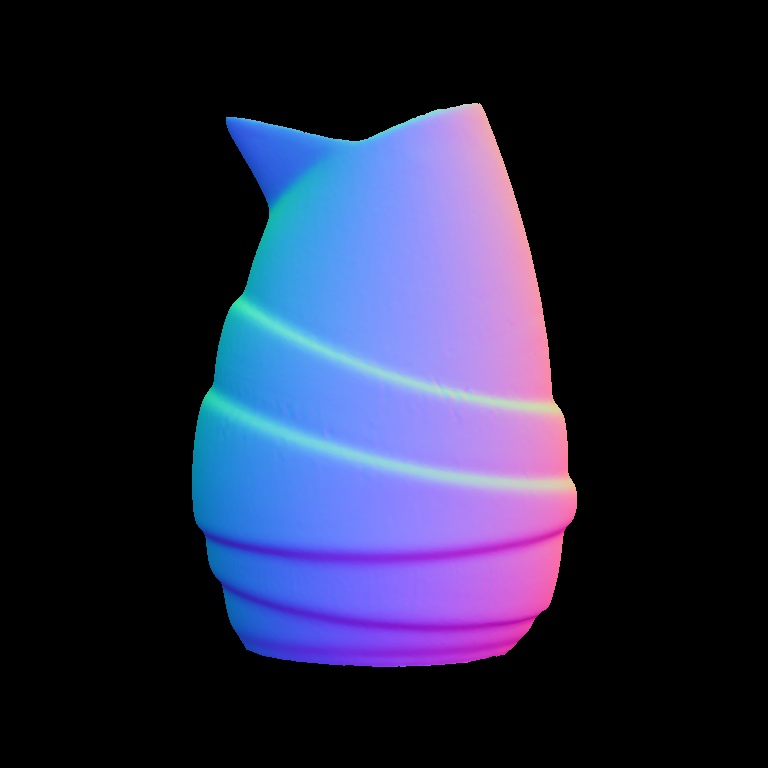}
    \\
    \includegraphics[width=\imagesize\textwidth]{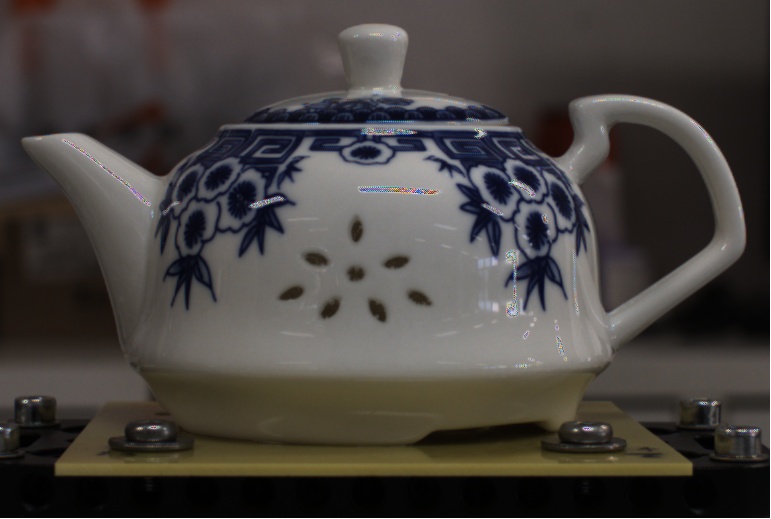}
    &
    \begin{overpic}[percent,width=\imagesize\textwidth]{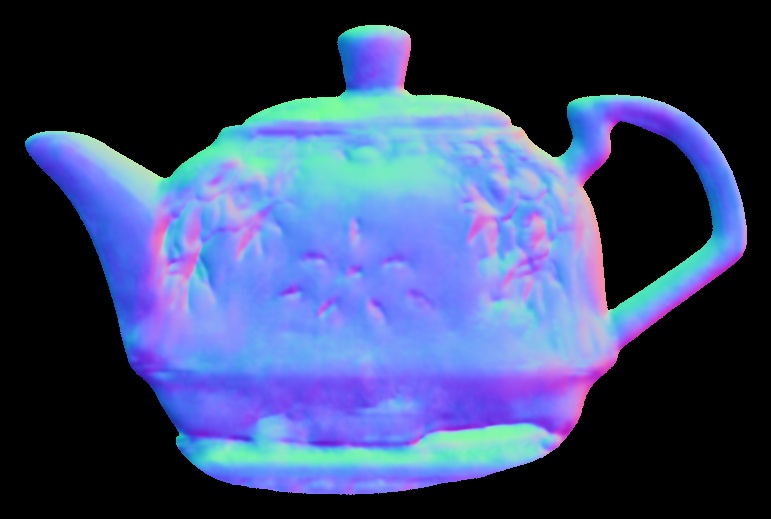}
    \put(80,4){\color{white}\scriptsize{27.32}}
    \end{overpic}
    &
    \includegraphics[width=\imagesize\textwidth]{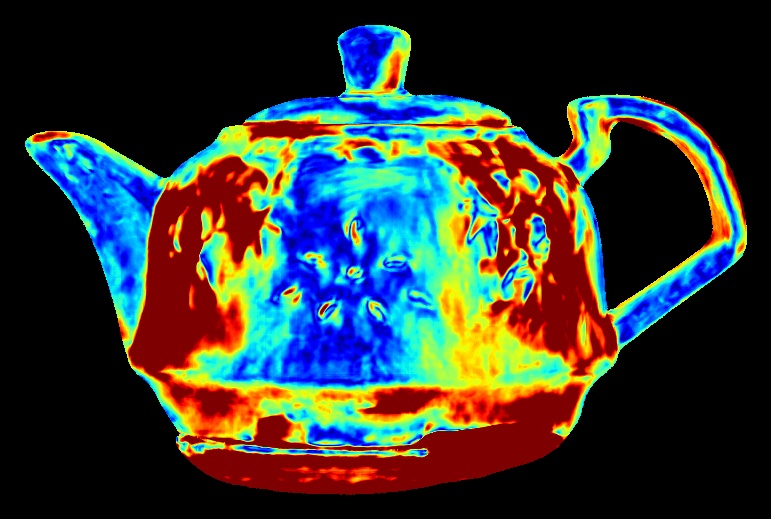}
    &
    \begin{overpic}[percent,width=\imagesize\textwidth]{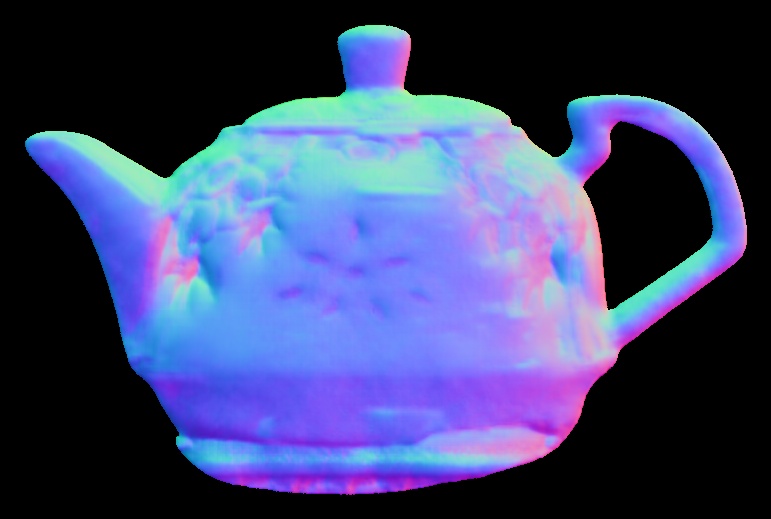}
    \put(80,4){\color{white}\scriptsize{21.68}}
    \end{overpic}
    &
    \includegraphics[width=\imagesize\textwidth]{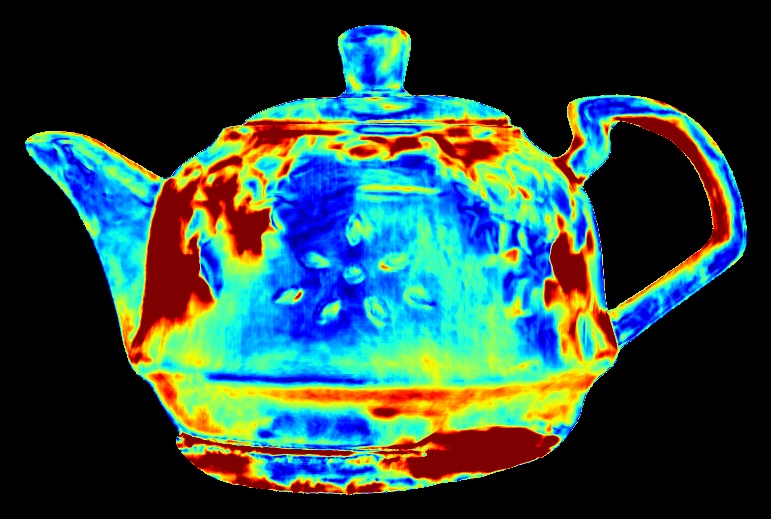}
    &
    \begin{overpic}[percent,width=\imagesize\textwidth]{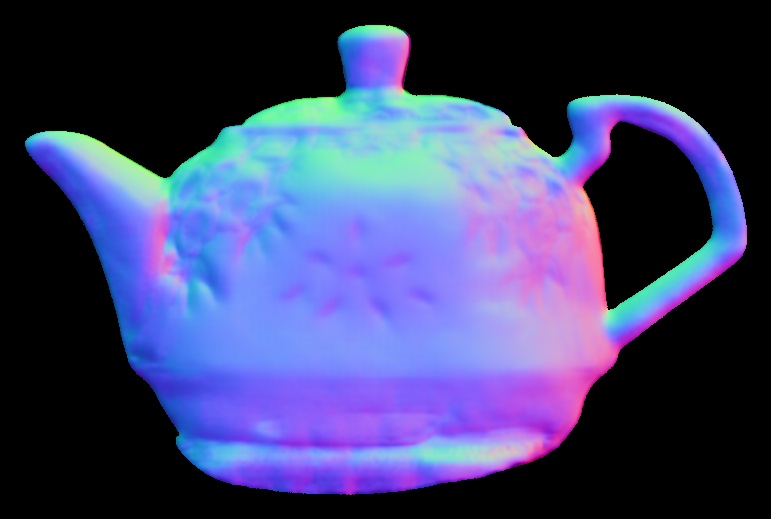}
    \put(80,4){\color{white}\scriptsize{18.97}}
    \end{overpic}
    &
    \includegraphics[width=\imagesize\textwidth]{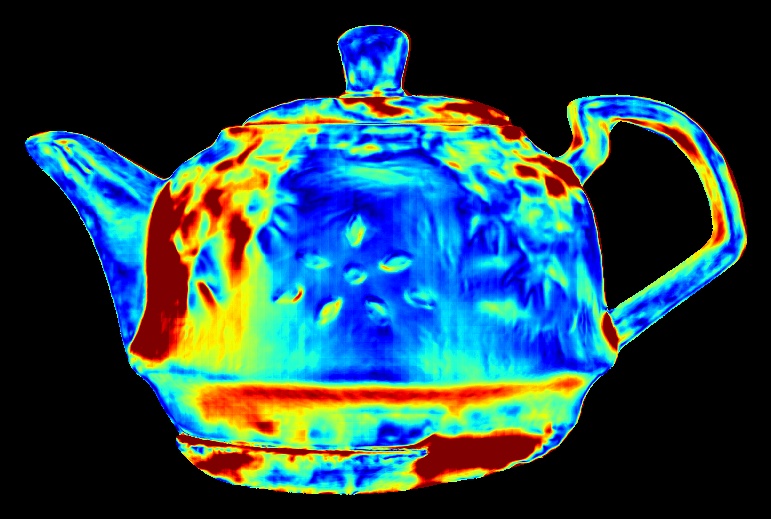}
    &
    \includegraphics[width=\imagesize\textwidth]{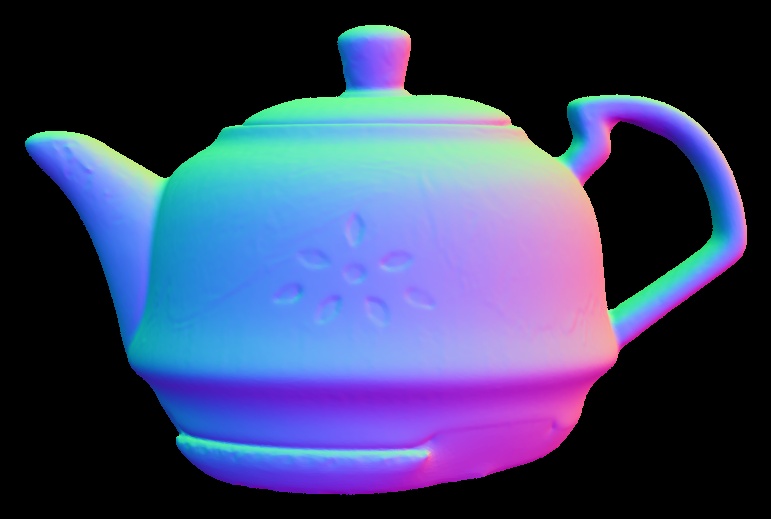}
    \\
    & Est.~Normal & AE Map & Est.~Normal & AE Map & Est.~Normal & AE Map &
\end{tabular}
}
}
    \vspace{-2mm}
    \caption{Evaluating \pa on single-view shape from polarization based on \deepsfp. ``AE Map'' denotes the angular error map between GT and estimated surface normal maps. The numbers at the bottom-right corner denote MAngE.}
    \label{fig:qual_eval_deepsfp}
\end{figure*}

\begin{table}[t]
    \caption{Comparisons of the single-view SfP method DeepSfP~\cite{ba19deepsfp} retrained on different datasets. We evaluate the retrained models on the original test set (``DP''), our captured dataset (``PN''), and the combination of the two datasets (``DP+PN'').}
    \vspace{-3mm}
    \begin{threeparttable}
\resizebox{\linewidth}{!}{
\begin{tabular}{c l c c c c c c}
\toprule
     \multirow{2}{*}{Test set} & \multirow{2}{*}{Training set} & \multicolumn{3}{c}{Angular error ($^\circ$) $\downarrow$} & \multicolumn{3}{c}{Accuracy (\%) $\uparrow$} \\
    && Mean & Median & RMSE & $\leq10^\circ$ & $\leq20^\circ$ & $\leq30^\circ$ \\
\midrule
\multirow{3}{*}{\rotatebox[origin=c]{0}{DP}} & \deepsfp\tnote{a} & 19.38 & \textbf{15.21} & 23.89 & \textbf{34.98} & 70.74 & 82.00 \\
& DeepSfP+MSO & \textbf{19.15} & 15.87 & \textbf{22.95} & 32.46 & 71.63 & \textbf{83.47} \\
& DeepSfP+PSO & 19.20 & 15.71 & 23.01 & 33.46 & \textbf{72.06} & 83.41 \\
\midrule
\multirow{3}{*}{\rotatebox[origin=c]{0}{PN}} & \deepsfp\tnote{a} & 30.69 & 26.88 & 35.86 & 9.80 & 34.49 & 58.26 \\
& DeepSfP+MSO & 24.22 & 21.14 & 28.42 & 12.95 & 47.88 & 72.26 \\
& DeepSfP+PSO & \textbf{22.93} & \textbf{20.08} & \textbf{27.58} & \textbf{14.39} & \textbf{50.08} & \textbf{77.47} \\
\midrule
\multirow{3}{*}{\rotatebox[origin=c]{0}{DP+PN}} & \deepsfp\tnote{a} & 22.21 & 18.13 & 26.88 & \textbf{24.68} & 57.39 & 76.07 \\
& DeepSfP+MSO & 20.42 & 17.19 & 24.31 & 23.23 & 61.32 & 80.67 \\
& DeepSfP+PSO & \textbf{20.13} & \textbf{16.81} & \textbf{24.15} & 24.47 & \textbf{62.00} & \textbf{81.92} \\
\bottomrule
\vspace{-2mm}
\end{tabular}
}
\vspace{-2mm}
\begin{tablenotes}[para]
\footnotesize{
\item[a] Original training data from \deepsfp
}
\end{tablenotes}
\end{threeparttable}

    \label{tab:quan_eval_deepsfp}
    \vspace{-3mm}
\end{table}

\paragraph{Evaluation on multiview SfP}
We choose the state-of-the-art multiview SfP method \pisr as the baseline. We adopt the test sample {\sc Shisa} from~\cite{nersp} for comparison, which comprises $6$ polarization images with GT camera poses and normal maps. 
To verify the effectiveness of \pa, we feed \pisr with the real-captured and generated polarization data, respectively, and the qualitative and quantitative results are shown in \fref{fig:eval_pisr}.
Chamfer Distance (CD) is adopted to evaluate the estimated meshes.
Our method can generate physically reasonable polarization properties (``Gen.~AoLP'' and ``Gen.~DoLP''). 
The reconstructed 3D shapes from generated images by \pa are comparable to those from real-captured images.  
More reconstruction results of \pisr are provided in the supplementary material.

\begin{figure*}[t]
    \footnotesize{
    \newcommand{\imagesize}{0.109}
\renewcommand{\arraystretch}{0.2}
\setlength{\tabcolsep}{1pt}{
\begin{tabular}{c c@{}c c@{}c c@{}c c@{}c}
    Input & \multicolumn{2}{c}{Diffusion on Polarized Images} & \multicolumn{2}{c}{Diffusion on AoLP\&DoLP} & \multicolumn{2}{c}{Diffusion on Enc.~AoLP\&DoLP} & \multicolumn{2}{c}{GT}
    \\
    \includegraphics[width=\imagesize\textwidth]{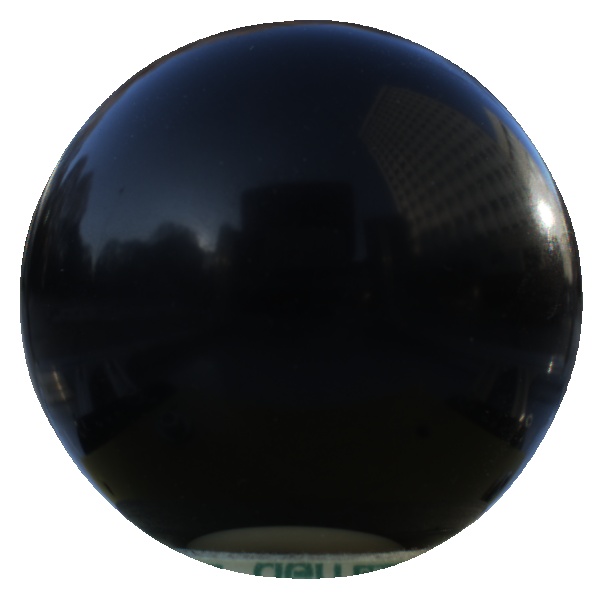}
    &
    \includegraphics[width=\imagesize\textwidth]{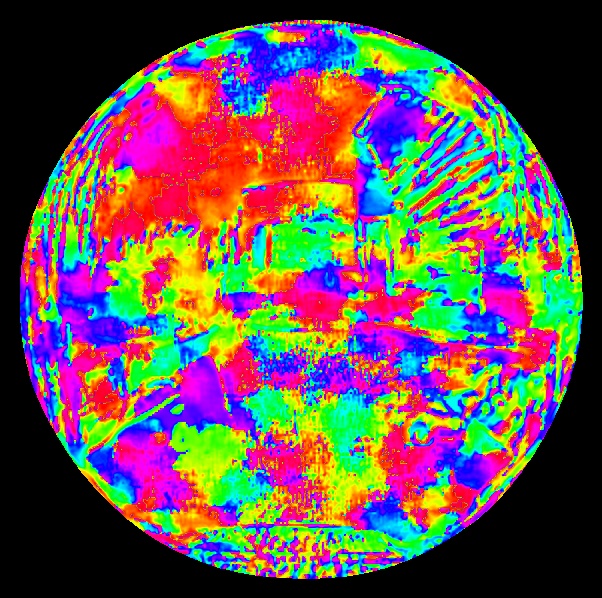}
    &
    \includegraphics[width=\imagesize\textwidth]{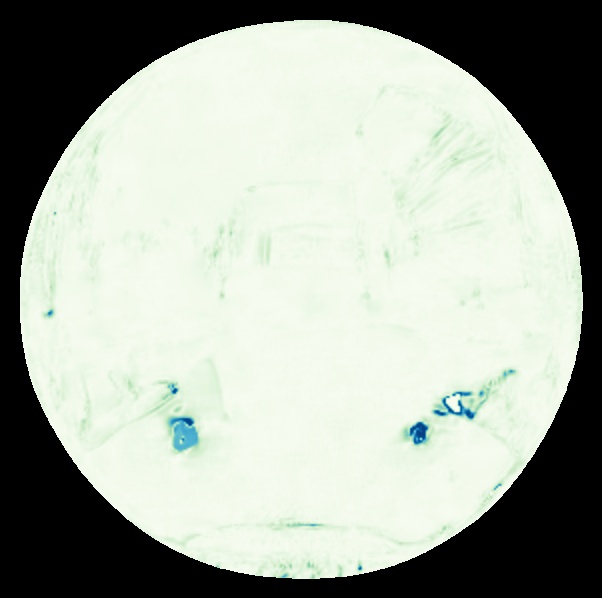}
    &
    \includegraphics[width=\imagesize\textwidth]{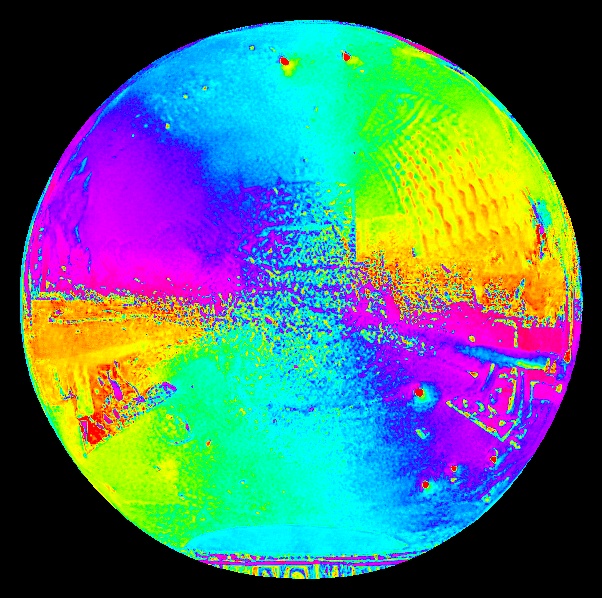}
    &
    \includegraphics[width=\imagesize\textwidth]{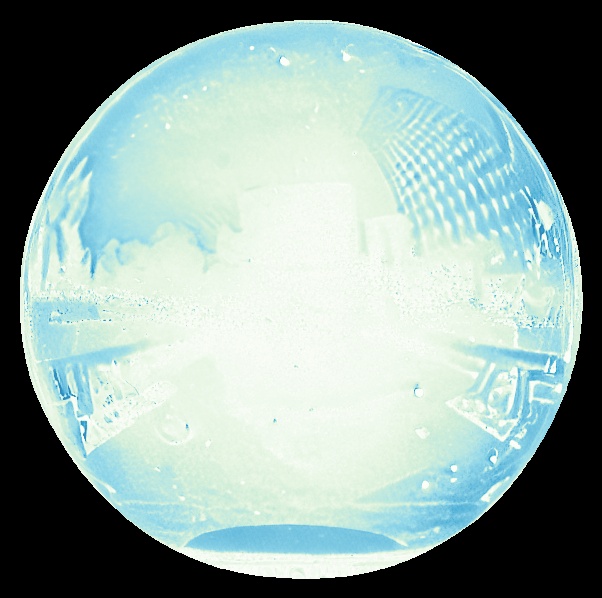}
    &
    \includegraphics[width=\imagesize\textwidth]{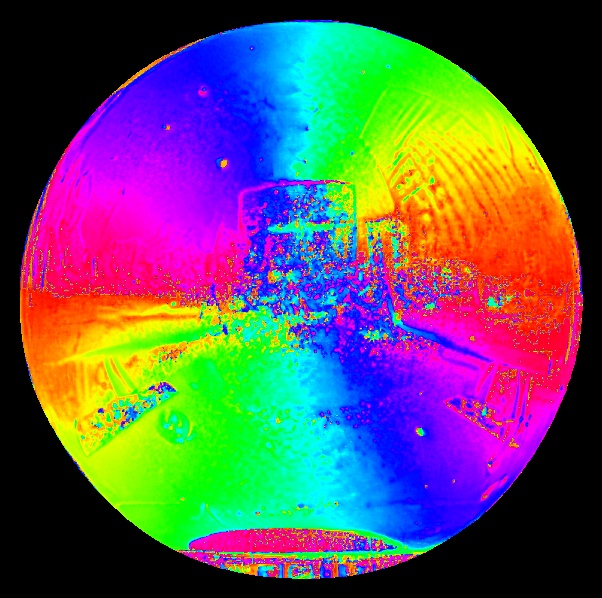}
    &
    \includegraphics[width=\imagesize\textwidth]{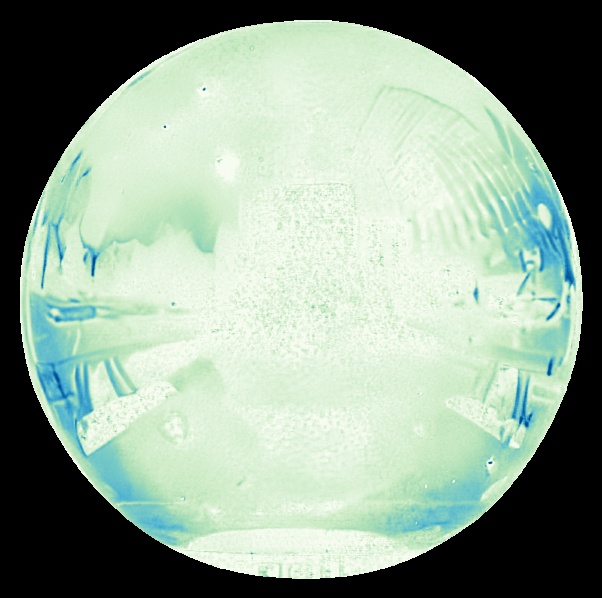}
    &
    \includegraphics[width=\imagesize\textwidth]{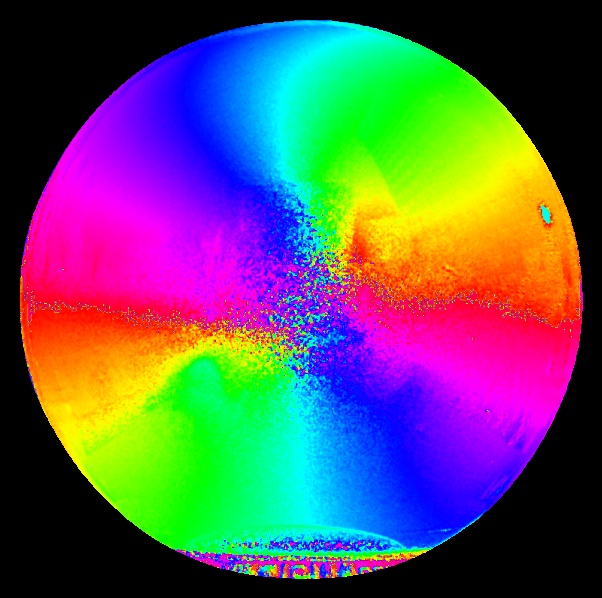}
    &
    \includegraphics[width=\imagesize\textwidth]{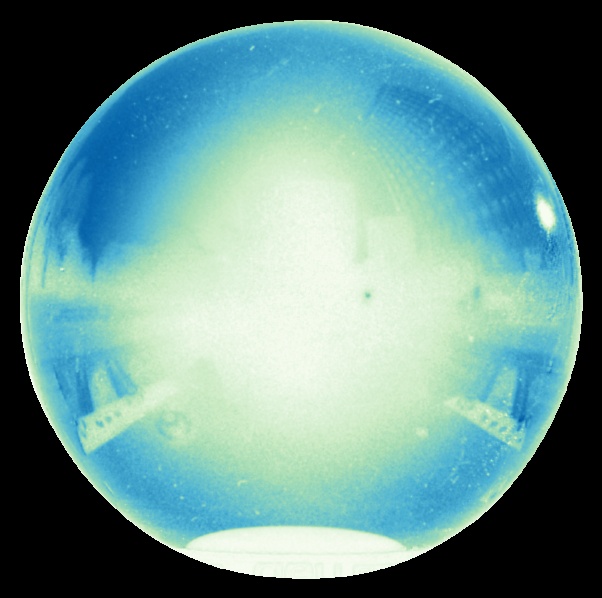}
    \\
    \includegraphics[width=\imagesize\textwidth]{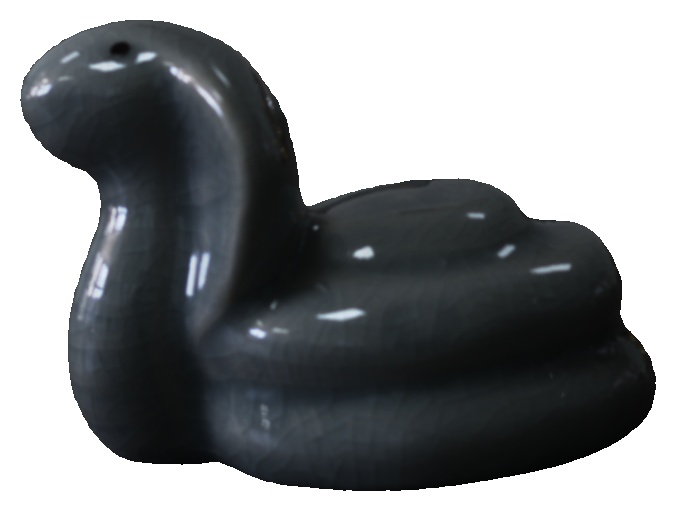}
    &
    \includegraphics[width=\imagesize\textwidth]{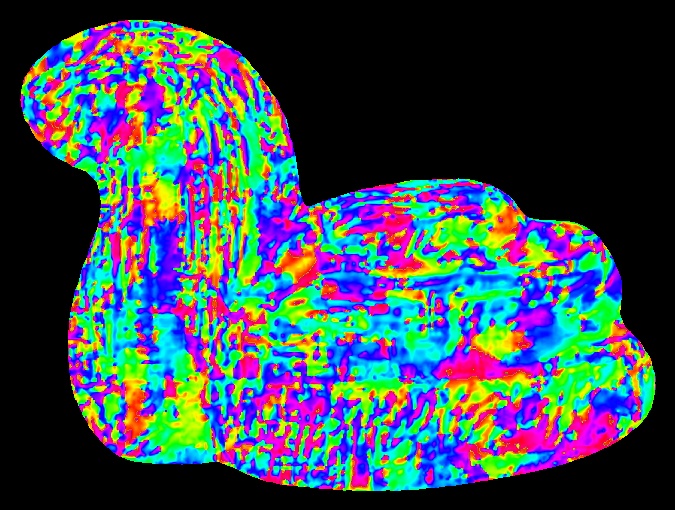}
    &
    \includegraphics[width=\imagesize\textwidth]{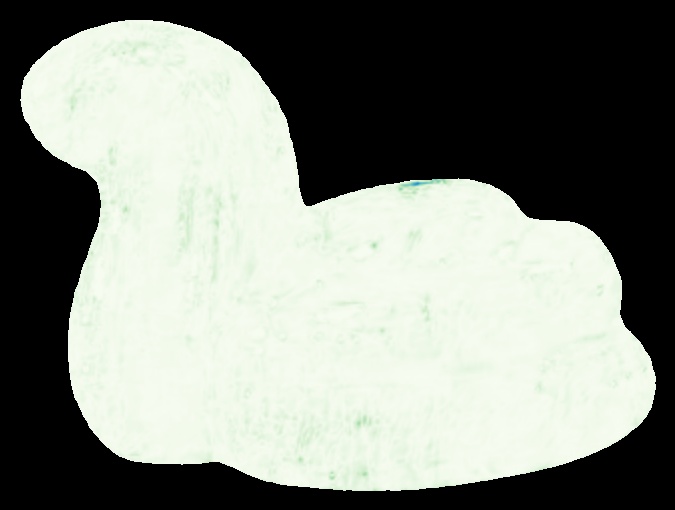}
    &
    \includegraphics[width=\imagesize\textwidth]{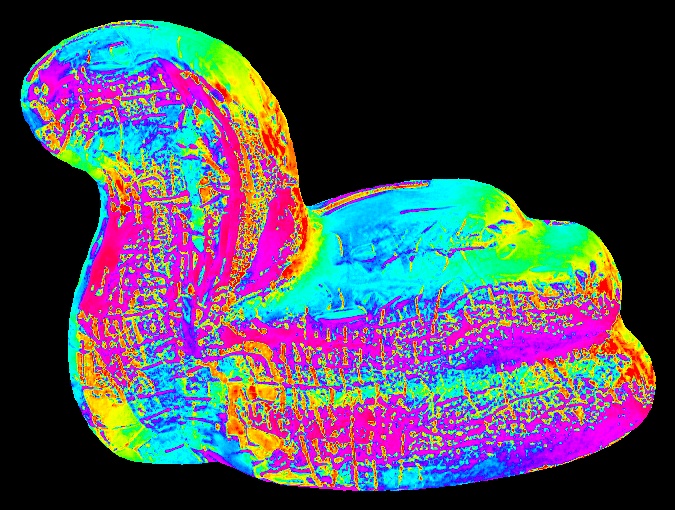}
    &
    \includegraphics[width=\imagesize\textwidth]{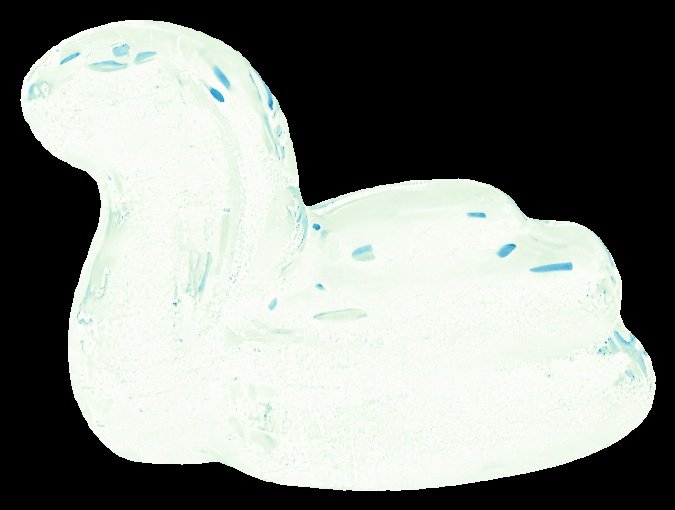}
    &
    \includegraphics[width=\imagesize\textwidth]{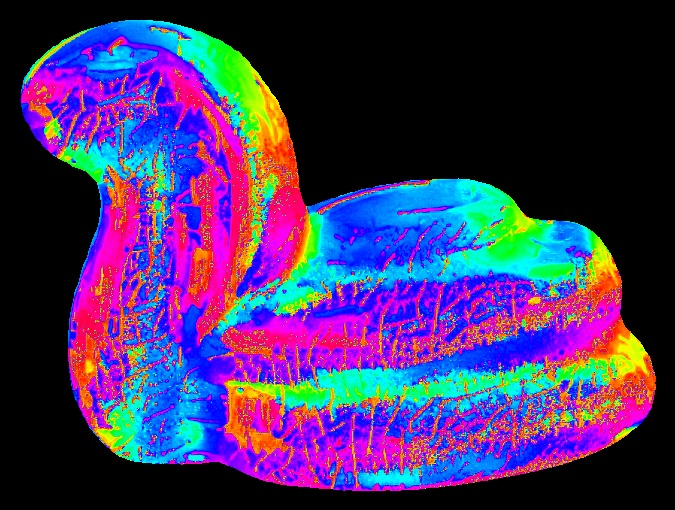}
    &
    \includegraphics[width=\imagesize\textwidth]{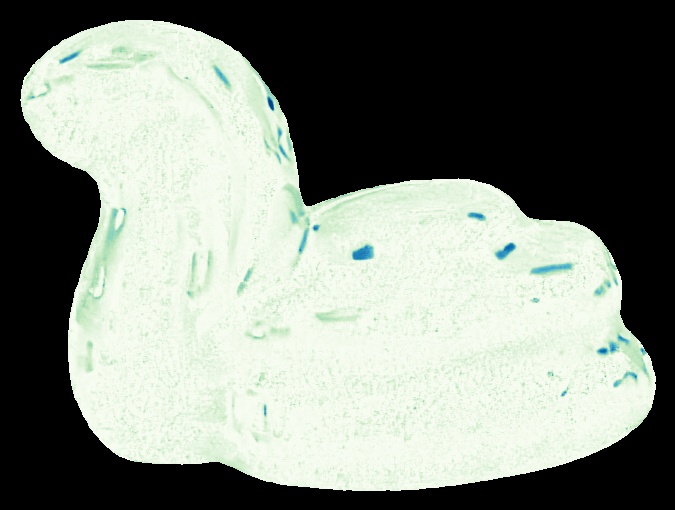}
    &
    \includegraphics[width=\imagesize\textwidth]{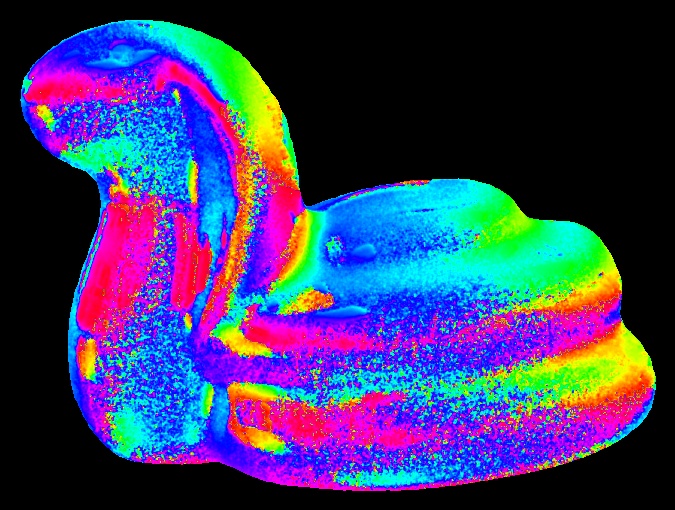}
    &
    \includegraphics[width=\imagesize\textwidth]{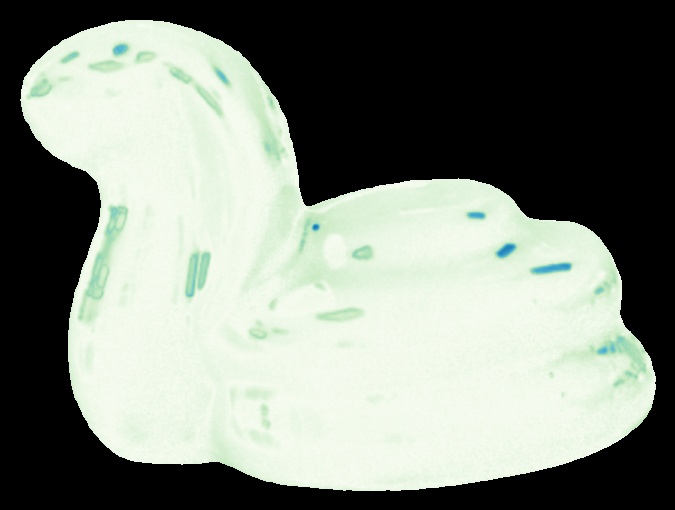}
    \\
    & AoLP & DoLP & AoLP & DoLP & AoLP & DoLP & AoLP & DoLP
    
\end{tabular}
}
    }
    \vspace{-2mm}
    \caption{Ablation study. Naively generating polarization images leads to unreliable polarization properties. \pa generates more accurate polarization predictions with encoded AoLP.}
    \label{fig:abl}
    \vspace{-3mm}
\end{figure*}

\paragraph{PolarStanford-ORB for enhancing single-view SfP}
\pa aims at easily producing polarization images from common RGB image datasets for learning-based tasks, mitigating the barrier of polarization data acquisition. To validate this, we randomly pick 300 RGB images from Stanford-ORB~\cite{kuang2023stanfordorb} and create PolarStanford-ORB (PSO) by \pa to expand the training dataset of \deepsfp, which contains only 236 training data. For a fair comparison, we also render 300 polarization images using Mitsuba with 3D assets from Stanford-ORB~\cite{kuang2023stanfordorb} named MitsubaStanford-ORB (MSO) for training set expansion. The DeepSfP model is retrained on the three different datasets, \ie, ``\deepsfp'', ``DeepSfP+MSO'', and ``DeepSfP+PSO'', with the same training strategy proposed in the original paper~\cite{ba19deepsfp}.

\begin{figure}[t]
    \centering
    \footnotesize{
    \newcommand{\imagesize}{0.078}
\renewcommand{\arraystretch}{0.2}
\setlength{\tabcolsep}{0.2pt}{
\begin{tabular}{c@{}c c@{}c c@{}c}
    \multicolumn{2}{c}{GT} & \multicolumn{2}{c}{\pa} & \multicolumn{2}{c}{\restormer}
    \\
    \includegraphics[width=\imagesize\textwidth]{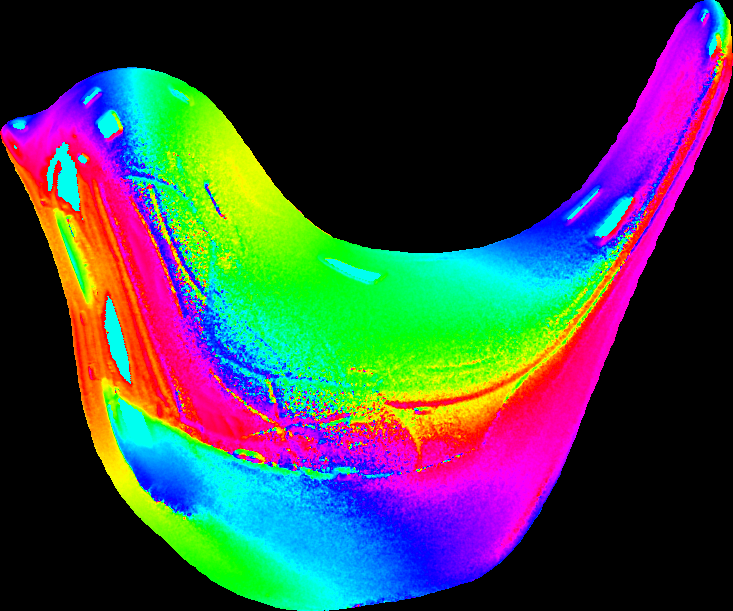}
    &
    \includegraphics[width=\imagesize\textwidth]
    {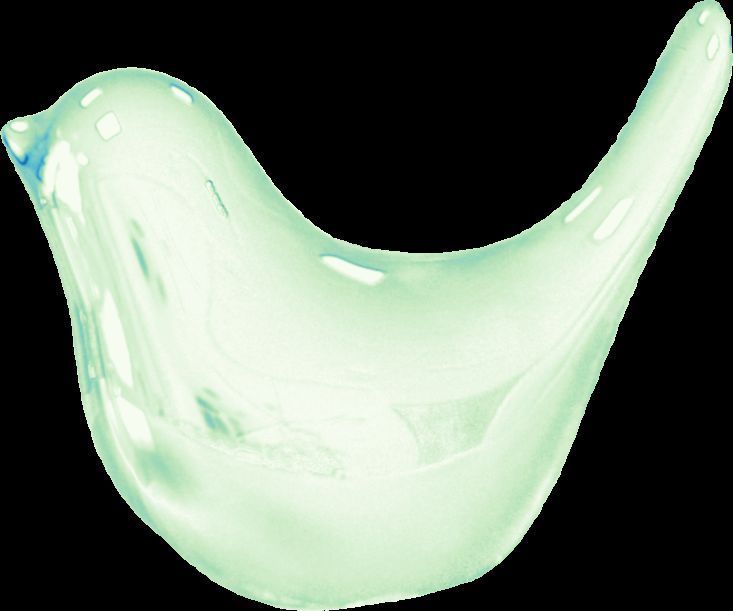}
    &
    \includegraphics[width=\imagesize\textwidth]{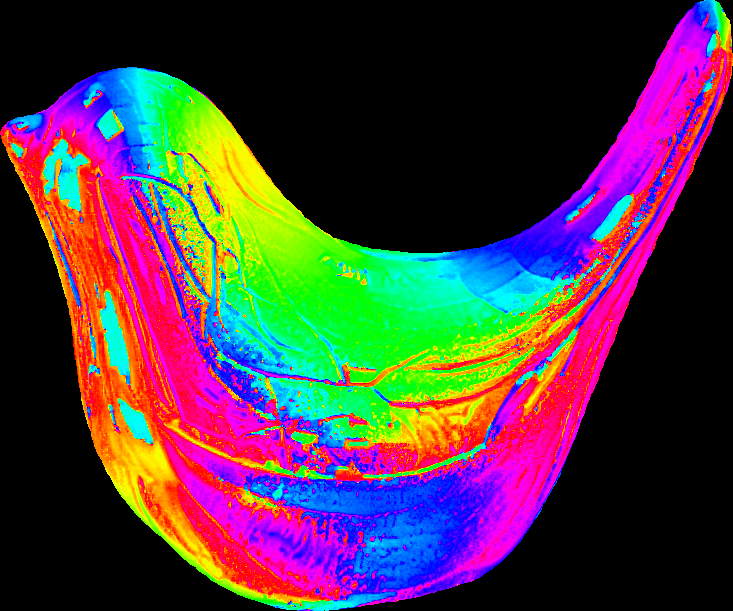}
    &
    \includegraphics[width=\imagesize\textwidth]{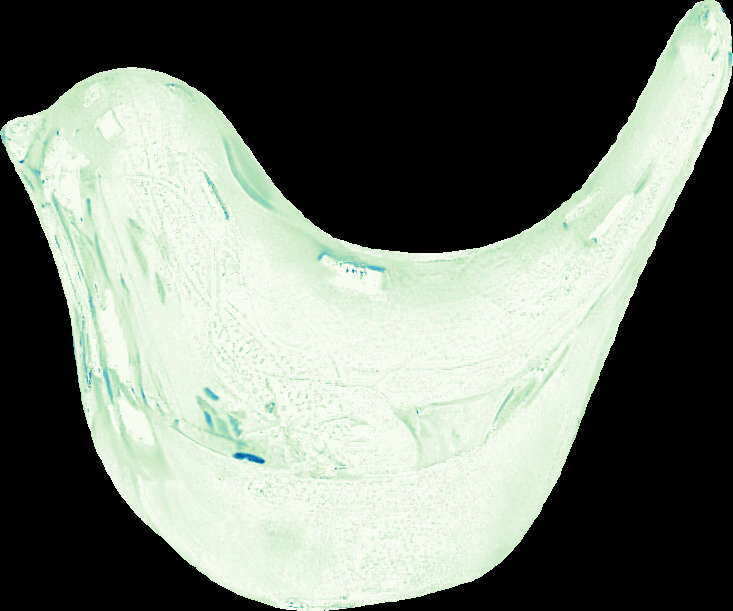}
    &
    \includegraphics[width=\imagesize\textwidth]{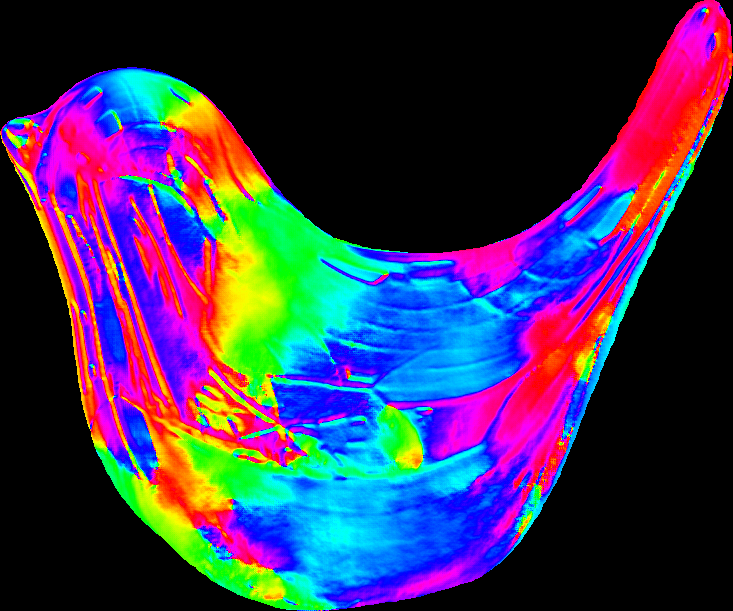}
    &
    \includegraphics[width=\imagesize\textwidth]{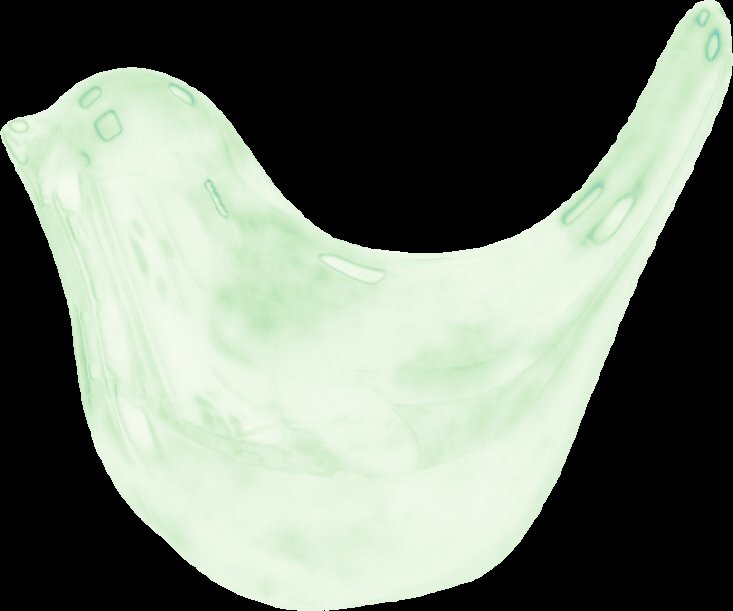}
    \\
    \includegraphics[width=\imagesize\textwidth]{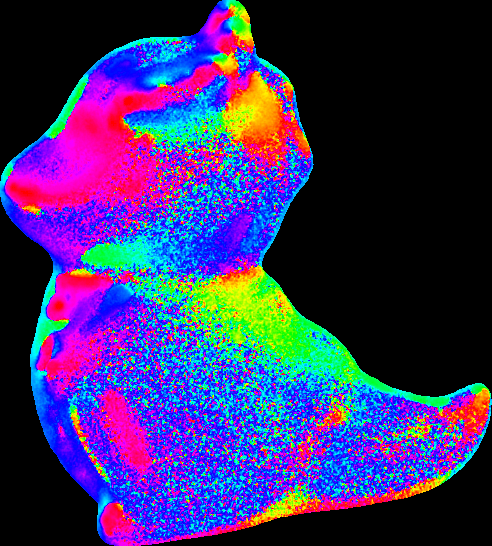}
    &
    \includegraphics[width=\imagesize\textwidth]{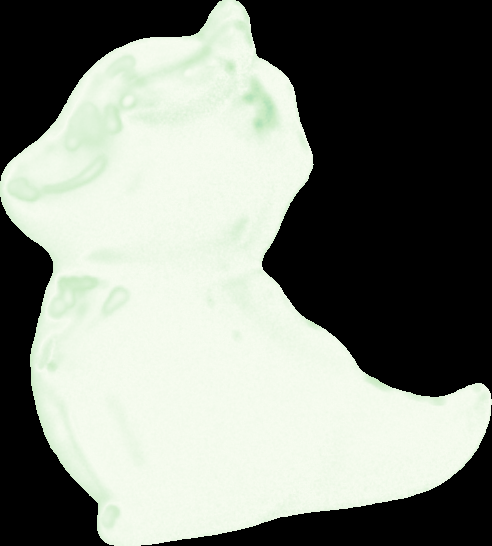}
    &
    \includegraphics[width=\imagesize\textwidth]{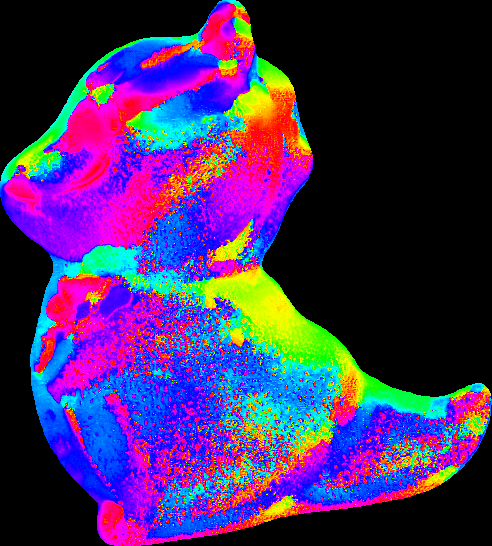}
    &
    \includegraphics[width=\imagesize\textwidth]{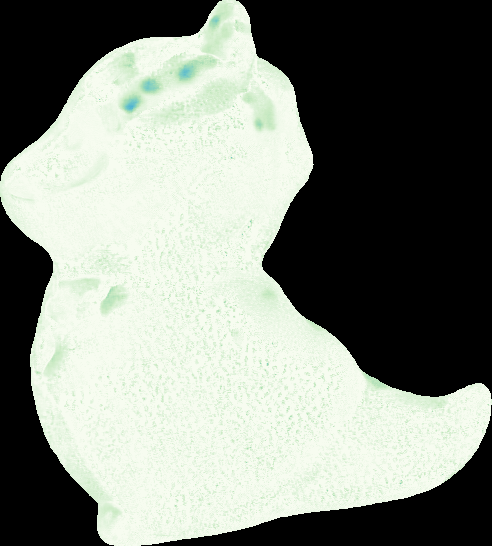}
    &
    \includegraphics[width=\imagesize\textwidth]{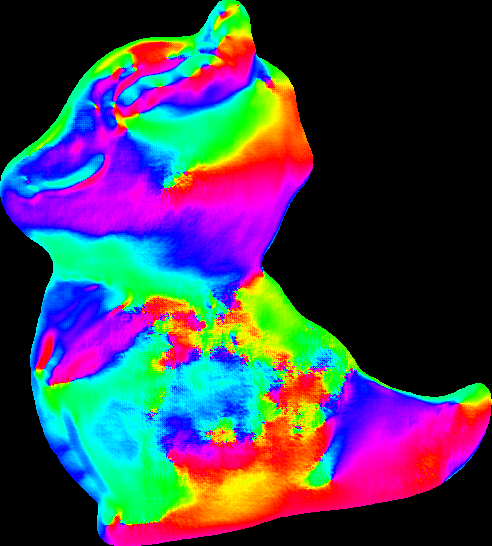}
    &
    \includegraphics[width=\imagesize\textwidth]{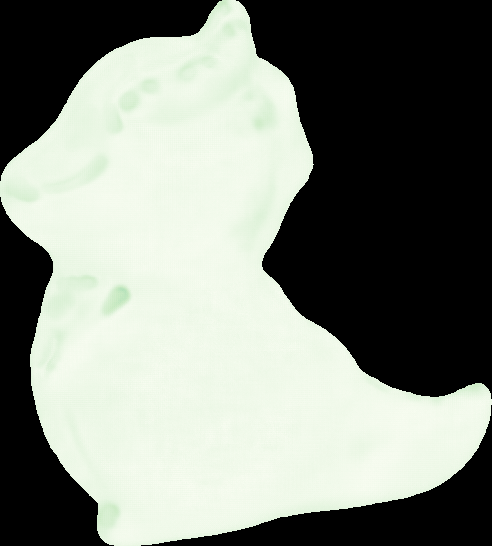}
    \\
     AoLP &  DoLP &  AoLP &  DoLP &  AoLP &  DoLP
    
\end{tabular}
}
    }
    \vspace{-2mm}
    \caption{Comparison of Transformer-based and Diffusion-based architectures for polarization synthesis.}
    \label{fig:r2}
    \vspace{-2mm}
\end{figure}

These three retrained models are evaluated on two test datasets: DP and PN, where ``DP'' refers to the real-world test set from \deepsfp containing 21 objects. Since the capture environment of ``DP'' is similar to the training dataset used in \deepsfp, we additionally capture a dataset named ``PN'' containing polarization images and surface normal pairs of $7$ objects to test the generalizability of the three retrained models. The capturing process of ``PN'' follows \diligent, and more details are provided in the supplementary material.
\Fref{fig:qual_eval_deepsfp} and \Tref{tab:quan_eval_deepsfp} present the qualitative and quantitative evaluations. The quantitative results for ``DeepSfP+MSO'' indicate that adding more polarization data enhances the model's generalization. When replacing the Mitsuba-rendered polarization data with \pa, we observe a similar reduction in MAngE for normal predictions, as shown in ``DeepSfP+PSO''. This suggests that polarization images from \pa can be effectively used in SfP. With PolarStanford-ORB augmentation, ``DeepSfP+PSO'' improves the Mean and RMSE metrics on the DeepSfP test set ``DP'', though the gains are marginal due to the similar distribution between the training and test sets. Testing on ``PN'', our newly captured real-world dataset with a different distribution, reveals more significant improvements, highlighting \pa's effectiveness in improving the generalizability.

\subsection{Ablation study}\label{sec:4.4}
To evaluate the effect of polarized information encoding in \pa, we test several diffusion targets against our polarization formulation: a) \pa generating 4 polarization images (\eg, $\mathbf{I}_{0^\circ},\mathbf{I}_{45^\circ},\mathbf{I}_{90^\circ},\mathbf{I}_{135^\circ}$). b) \pa generating AoLP and DoLP maps. c) \pa generating encoded AoLP and DoLP maps. 
For setting a), we generate the four polarization images concatenated in the height and width dimensions as a 2$\times$2 grid. For b), we also use a single denoising U-Net to produce AoLP and DoLP maps as two channels in the output. All the ablative experiments are conducted with the same training dataset and training strategy as \pa.
As shown in \fref{fig:abl} and \Tref{tab:quan_abl}, choice a) predicting polarization images only retains RGB radiance information but hardly recovers polarization properties, due to the absence of physical constraint on polarization. Both quantitative and qualitative results show that choice b) produces worse results than our method. Since the range of AoLP values is within $[-90^\circ, 90^\circ]$, directly regressing AoLP values can disrupt its inherent periodicity making the network challenging to learn.

\begin{table}[t]
    \caption{Ablation study of each polarization representation in \pa on our real-world dataset.}
    \vspace{-3mm}
    \footnotesize{
\begin{tabular}{lcccc}
    \toprule
    Generation Target & PSNR$\uparrow$ & SSIM$\uparrow$ & MAngE$\downarrow$ & MAbsE$\downarrow$ \\
    \midrule
    Polarization images & 23.23 & 0.9165 & 45.67 & 0.1233 \\
    AoLP\&DoLP & 40.57 & 0.9904 & 29.46 & 0.1100 \\
    Encoded AoLP\&DoLP & \textbf{41.74} & \textbf{0.9927} & \textbf{25.33} & \textbf{0.1075}  \\
    \bottomrule
    \vspace{-7mm}
\end{tabular}
    }
    \label{tab:quan_abl}
\end{table}

We also test different network architectures trained on the same training dataset to validate our motivation for employing diffusion models in polarization simulation. The synthesis results using \restormer as the baseline are shown in \fref{fig:r2}. Compared to \pa, \restormer struggles with AoLP and DoLP estimation due to its reliance on large-scale datasets, and the scarcity of polarization data limits its effectiveness in estimating polarization properties. In contrast, our chosen diffusion models, with their strong zero-shot capabilities, demonstrate superior performance in generating polarization images from a single RGB input.

\section{Conclusion}
We propose \pa, a novel diffusion-based polarization image generation framework that generates full polarization properties. Compared to polarization image renderer \mitsuba, \pa only takes as input a single RGB image rather than complex PBR components, which greatly reduces the barrier of large-scale polarization dataset creation. Extensive experiments show that \pa generates photorealistic images while faithfully recovering physical polarization information. As demonstrated in the supplementary material, the synthesized data supports downstream tasks such as shape-from-polarization, glass segmentation, and reflection control.

\section*{Acknowledgement}
This work was supported by Hebei Natural Science Foundation Project No. 242Q0101Z, Beijing-Tianjin-Hebei Basic Research Funding Program No. F2024502017, National Natural Science Foundation of China (Grant No. 62472044, U24B20155, 62225601, U23B2052, 62136001), Beijing Natural Science Foundation Project No. L242025. We thank openbayes.com for providing computing resource.
{
    \small
    \bibliographystyle{lib/ieeenat_fullname}
    \bibliography{PolarAnything}
}
\end{document}